\documentclass[10pt,twocolumn,letterpaper]{article}

\usepackage[pagenumbers]{cvpr} 
\usepackage{graphicx}
\usepackage{amsmath}
\usepackage{amssymb}
\usepackage{amsthm}
\usepackage{bm}
\usepackage{booktabs}
\usepackage{csquotes}
\usepackage{array}
\usepackage{xcolor, colortbl}
\makeatletter
\@namedef{ver@everyshi.sty}{}
\makeatother
\usepackage{tikz}
\usetikzlibrary{shapes}
\usepackage[pagebackref,breaklinks,colorlinks]{hyperref}
\usepackage[capitalize]{cleveref}
\crefname{section}{Sec.}{Secs.}
\Crefname{section}{Section}{Sections}
\Crefname{table}{Table}{Tables}
\crefname{table}{Tab.}{Tabs.}

\newcommand{\x}{\bm{x}}
\newcommand{\w}{\bm{w}}
\renewcommand{\L}{\mathcal{L}}
\newcommand{\R}{\mathbb{R}}
\newcommand{\I}{\mathbb{I}}
\newcommand{\norm}[1]{|\!| #1 |\!|}
\newcommand{\diff}[2]{\frac{\mathrm{d} #1}{\mathrm{d} #2}}
\newcommand{\dell}[2]{\frac{\partial #1}{\partial #2}}
\newcommand{\mean}{\mathrm{mean}}
\newcommand{\std}{\mathrm{std}}
\newcommand{\MSE}{\mathrm{MSE}}
\newcommand{\CE}{\mathrm{CE}}
\newcommand{\BCE}{\mathrm{BCE}}
\newcommand{\smL}{\mathrm{sm}L^1}
\newcommand{\N}{\mathbb{N}}
\newcommand{\e}[1]{\mathrm{e}^{#1}}
\newcommand{\sfx}{\mathsf{x}}
\newcommand{\sfy}{\mathsf{y}}
\newcommand{\sfw}{\mathsf{w}}
\newcommand{\sfh}{\mathsf{h}}
\newcommand{\obj}{\mathrm{obj}}
\newcommand{\noobj}{\mathrm{noobj}}
\newcommand{\out}{\mathrm{out}}

\newcommand{\map}{\mathit{mAP}}
\newcommand{\ap}{\mathit{AP}}
\newcommand{\iou}{\mathit{IoU}}
\newcommand{\auroc}{\mathit{AuROC}}
\newcommand{\ACE}{\mathit{ACE}}
\newcommand{\MCE}{\mathit{MCE}}
\newcommand{\ECE}{\mathit{ECE}}
\newcommand{\acc}{\mathrm{acc}}
\newcommand{\conf}{\mathrm{conf}}

\DeclareMathOperator{\argmax}{arg\,max}

\definecolor{LightCyan}{rgb}{0.88,1,1}
\definecolor{LightGray}{gray}{0.92}
\definecolor{silver}{rgb}{0.75, 0.75, 0.75}
\definecolor{ao}{rgb}{0.0, 0.0, 1.0}
\definecolor{ao(english)}{rgb}{0.0, 0.5, 0.0}

\newtheorem{thm}{Theorem}
\newtheorem*{thm-non}{Theorem 1}

\begin{document}

\title{Gradient-Based Quantification of Epistemic Uncertainty \\ for Deep Object Detectors}

\author{Tobias Riedlinger
\qquad Matthias Rottmann
\qquad Marius Schubert
\qquad Hanno Gottschalk\\\\
School of Mathematics and Natural Sciences\\
University of Wuppertal\\
{\tt\small $\{$riedlinger, rottmann, mschubert, hgottsch$\}$@uni-wuppertal.de}

}
\maketitle

\begin{abstract}
        The vast majority of uncertainty quantification methods for deep object detectors such as variational inference are based on the network output.
        Here, we study gradient-based epistemic uncertainty metrics for deep object detectors to obtain reliable confidence estimates.
        We show that they contain predictive information and that they capture information orthogonal to that of common, output-based uncertainty estimation methods like Monte-Carlo dropout and deep ensembles.
        To this end, we use meta classification and meta regression to produce confidence estimates using gradient metrics and other baselines for uncertainty quantification which are in principle applicable to any object detection architecture.
        Specifically, we employ false positive detection and prediction of localization quality to investigate uncertainty content of our metrics and compute the calibration errors of meta classifiers.
        Moreover, we use them as a post-processing filter mechanism to the object detection pipeline and compare object detection performance.
        Our results show that gradient-based uncertainty is itself on par with output-based methods across different detectors and datasets.
        More significantly, combined meta classifiers based on gradient and output-based metrics outperform the standalone models.
        Based on this result, we conclude that gradient uncertainty adds orthogonal information to output-based methods.
        This suggests that variational inference may be supplemented by gradient-based uncertainty to obtain improved confidence measures, contributing to down-stream applications of deep object detectors and improving their probabilistic reliability.
\end{abstract}

\section{Introduction}
\label{sec: intro}
\begin{figure}
    \centering
    \resizebox{\linewidth}{!}{
    \begin{tikzpicture}
    \node (top_img) at (0, 2) {\includegraphics[width=\linewidth]{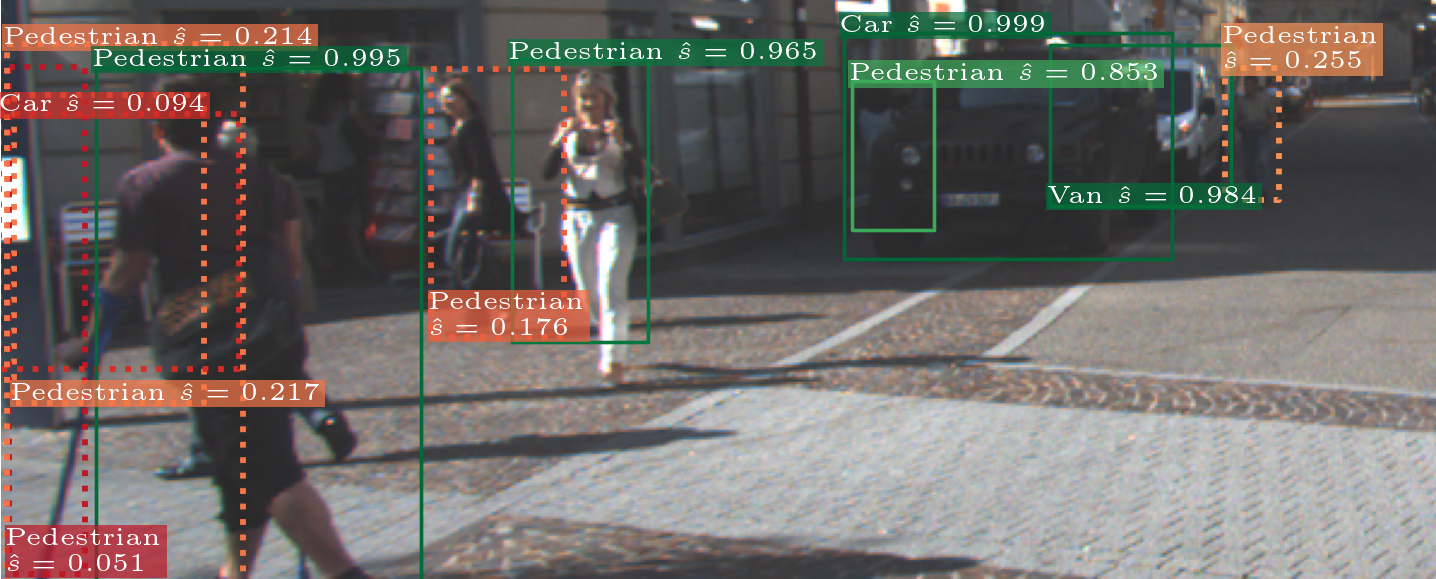}};
    \node (bottom_img) at (0, -1.4) {\includegraphics[width=\linewidth]{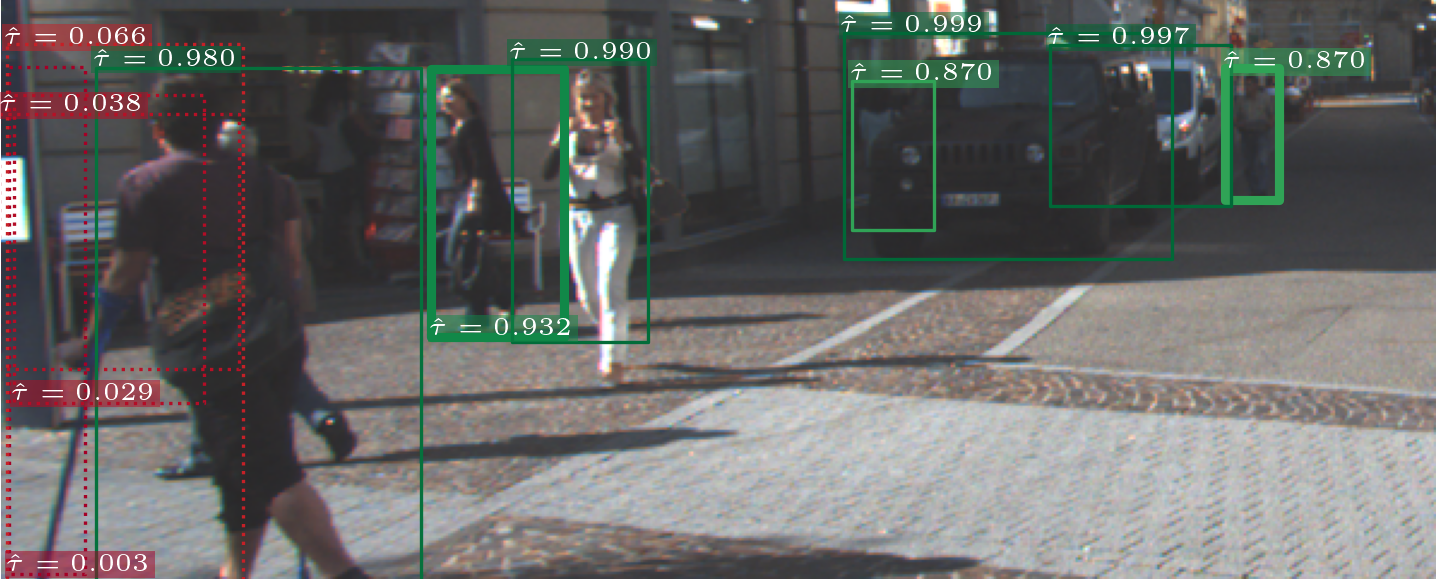}};
    \draw (-4.3, -3.2) rectangle (4.3, 3.8);
    \node[fill=white, inner sep=2pt] at (3, 0.7) {\color{blue} DNN Score $\hat{s}$};
    \node[fill=white, inner sep=2pt] at (2.4, -2.7) {\color{blue} Gradient confidence $\hat{\tau}$};
    \end{tikzpicture}
    }
    \caption{Object detection in a street scene (full images in \cref{app: results}). Top coloration: Score $\hat{s}$; bottom coloration: instance-wise gradient-based confidence $\hat{\tau}$ obtained by our method.
    Dashed boxes here indicate the discarding at any confidence threshold in $[0.3, 0.85]$.
    The top image contains FNs which are not separable from correctly discarded boxes based on the score (lower threshold would lead to FPs).
    In the bottom image, those $\hat{s}$-FNs are assigned higher confidences and there is a large range of thresholds with no FPs.
    }
    \label{fig: reliability classifiers}
\end{figure}
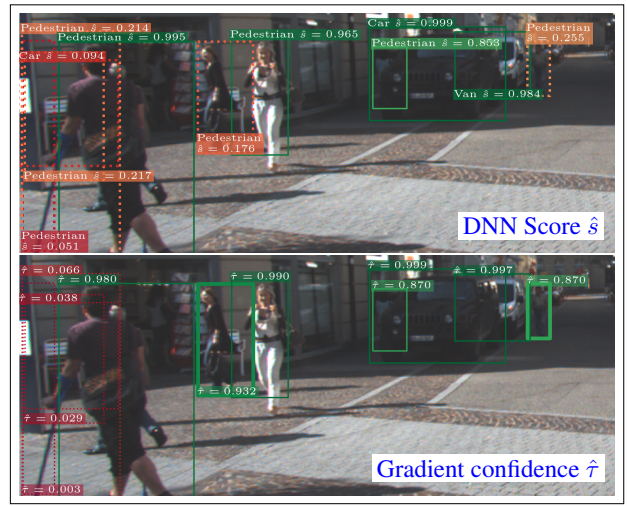
Deep artificial neural networks (DNNs) designed for tasks such as object detection or semantic segmentation provide a probabilistic prediction on given feature data such as camera images.
Modern deep object detection architectures \cite{liu2016ssd, redmon2018yolov3, ren2015faster, lin2017focal,cai2018cascade} predict bounding boxes for instances of a set of learned classes on an input image.

The so-called \emph{objectness} or \emph{confidence score} indicates the probability of the existence of an object for each predicted bounding box.
Throughout this work, we will refer to this quantity which the DNN learns as the \enquote{score}.
For applications of deep object detectors such as automated surgery or driving, the reliability of this component is crucial.
See, for example the detection in the top panel of \cref{fig: reliability classifiers} where each box is colored from red (low score) to green (high score).
Apart from the accurate, green boxes, boxes with a score below $0.3$ (dashed) contain true and false predictions which \emph{cannot be reliably separated in terms of their score}.
In addition, it is well-known that DNNs tend to give mis-calibrated scores \cite{szegedy2013intriguing, goodfellow2014explaining, guo2017calibration} that are oftentimes over-confident and may also lead to unreliable predictions.
Over-confident predictions might render an autonomous driving system inoperable by perceiving non-existent instances (false positives / FP).
Perhaps even more detrimental, under-confidence may lead to overlooked (false negative / FN) predictions possibly endangering humans outside autonomous vehicles like pedestrians and cyclists, as well as passengers.

Apart from modifying and improving the detection architecture or the loss function, there exist methods to estimate prediction confidence more involved than the score in order to remedy these issues\cite{miller2018dropout,lyu2020probabilistic,schubert2020metadetect}.
We use the term \enquote{confidence} more broadly than \enquote{score} to refer to quantities which represent the probability of a detection being correct.
Such a quantity should reflect the model's overall level of competency when confronted with a given input and is intimately linked to prediction uncertainty.
Uncertainty for statistical models, in particular DNNs, can broadly be divided into two types \cite{hullermeier2021aleatoric} depending on their primary source \cite{gal2016uncertainty,kendall2017uncertainties}.
Whereas aleatoric uncertainty is mainly founded in the stochastic nature of the data generating process, epistemic uncertainty stems from the probabilistic nature of sampling data for training, as well, as the choice of model and the training algorithm. 
The latter is technically reducible by obtaining additional training data and is the central subject of our method.

Due to the instance-based nature of deep object detection, modern ways of capturing epistemic uncertainty are mainly based on the instance-wise DNN output.
From a theoretical point of view, Bayesian DNNs \cite{denker1990transforming, mackay1992practical} represent an attractive framework for capturing epistemic uncertainty for DNNs by modeling their weights as random variables.
Practically, this approach introduces a large computational overhead making its application infeasible for object detection.
Therefore, in variational inference approaches, weights are sampled from predefined distributions to address this.
These famously include methods like Monte-Carlo (MC) dropout \cite{srivastava2014dropout, gal2016dropout} generating prediction variance by performing several forward passes under active dropout.
The same idea underlies deep ensemble sampling \cite{lakshminarayanan2016simple} where separately trained models with the same architecture produce variational forward passes.
Other methods relying only on the classification output per instance can also be applied to object detection such as softmax entropy or energy score methods.

A number of other, strong uncertainty quantification methods that do not only rely on the classification output has also been developed for image classification architectures \cite{corbiere2019addressing,malinin2018predictive,oberdiek2018classification,ramalho2020density}.
However, the \emph{transfer of such methods to object detection} frameworks can pose serious challenges, if at all possible, due to architectural restrictions.
For example, the usage of a learning gradient evaluated at the network's own prediction was proposed \cite{oberdiek2018classification} to contain epistemic uncertainty information for image classification and investigated for out-of-distribution (OoD) data.
The method has also been applied gainfully to natural language understanding \cite{vasudevan2019towards} where gradient metrics and deep ensemble uncertainty were aggregated to obtain well-calibrated and predictive confidence measures on OoD data.
The epistemic content of gradient uncertainty has further been explored in \cite{huang2021importance} in the classification setting by observing shifts in the data distribution.

We propose a way to compute gradient features for the prediction of deep object detectors.
We show that they perform on par with state-of-the-art uncertainty quantification methods and that they contain information that can not be obtained from output- or sampling-based methods.
In particular, \emph{we summarize our main contributions as follows}:
\begin{itemize}
\setlength{\itemsep}{0pt}
    \item We introduce a way of generating gradient-based uncertainty metrics for modern object detection architectures, allowing to generate uncertainty information from hidden network layers.
    \item We investigate the performance of gradient metrics in terms of meta classification (FP detection), calibration and meta regression (prediction of intersection over union \(\iou\) with the ground truth) and compare them to other means to quantify/approximate epistemic uncertainty and investigate mutual redundancy as well as detection performance through gradient uncertainty.
    \item We explicitly investigate the FP/FN-tradeoff for pedestrian detection based on the score and meta classifiers.
    \item We provide a theoretical treatment of the computational complexity of gradient metrics in comparison with MC dropout and deep ensembles and show that their FLOP count is similar at worst.
\end{itemize}
An implementation of our method will be made publicly available at \href{https://github.com/tobiasriedlinger/gradient-metrics-od}{https://github.com/tobiasriedlinger/gradient-metrics-od}.
A video illustration of our method is publicly available at \href{https://youtu.be/L4oVNQAGiBc}{https://youtu.be/L4oVNQAGiBc}.

\section{Related work}
\label{sec: related work}
\textbf{Epistemic uncertainty for deep object detection.}
Sampling-based uncertainty quantification such as MC dropout and deep ensembles have been investigated in the context of object detection by several authors in the past \cite{harakeh2020bayesod,miller2018dropout, miller2019merging, miller2019benchmarking, lyu2020probabilistic}.
They are straight-forward to implement into any architecture and yield output variance for all bounding box features.

Harakeh \etal \cite{harakeh2020bayesod} employed MC dropout and Bayesian inference as a replacement of Non-Maximum Suppression (NMS) to get a joint estimation of epistemic and aleatoric uncertainty.

Similarly, uncertainty measures of epistemic kind were obtained by Kraus and Dietmayer \cite{kraus2019uncertainty} from MC dropout.
Miller \etal \cite{miller2018dropout} investigated MC dropout as a means to improve object detection performance in open-set conditions.
Different merging strategies for samples from MC dropout were investigated by Miller \etal \cite{miller2019merging} and compared with the influence of merging boxes in deep ensembles of object detectors \cite{miller2019benchmarking}.
It was found that even for small ensemble sizes, deep ensembles outperform MC dropout sampling.
Lyu \etal \cite{lyu2020probabilistic} aggregated deep ensemble samples as if produced from a single detector to obtain improved detection performance.
A variety of uncertainty measures generated from proposal box variance pre-NMS called MetaDetect was investigated by Schubert \etal \cite{schubert2020metadetect}.
In generating advanced scores and $\iou$ estimates, it was reported that the obtained information is largely redundant with MC dropout uncertainty features but less computationally demanding.
All of the above methods are based on the network output and generate variance in by aggregating prediction proposals in some manner.
Moreover, a large amount of uncertainty quantification methods based on classification outputs can be directly applied to object detection \cite{hendrycks2016baseline,liu2020energy}.
Little is known about other methods developed for image classification that are not directly transferable to object detection due to architectural constraints, such as activation-based \cite{corbiere2019addressing} or gradient-based \cite{oberdiek2018classification} uncertainty.
The central difficulty in such an application lies in the fact that different predicted instances depend on shared latent features or DNN weights such that the base method can only estimate uncertainty for the entire prediction (consisting of all instances) instead of individual uncertainties for each instance.
We show that gradient uncertainty information can be extracted from hidden layers in object detectors.
We seek to determine how they compare against output-based methods and show that they contain orthogonal information.

\textbf{Meta classification and meta regression.}
The term meta classification refers to the discrimination of TPs from FPs on the basis of uncertainty metrics which was first explored by Hendrycks and Gimpel \cite{hendrycks2016baseline} to detect OoD samples based on the maximum softmax probability.

Since then, the approach has been applied to natural language processing \cite{vasudevan2019towards}, semantic segmentation\cite{chan2019metafusion, maag2020time, rottmann2019detection, MetaSeg, rottmann2019uncertainty}, instance segmentation in videos \cite{maag2020improving} and object detection \cite{schubert2020metadetect, kowol2020yodar} to detect FP predictions on the basis of uncertainty features accessible during inference.
Moreover, meta regression (the estimation of $\iou$ based on uncertainty in the same manner) was also investigated \cite{maag2020time,maag2020improving,rottmann2019detection,rottmann2019uncertainty, schubert2020metadetect} showing large correlations between estimates and the true localization quality. 

Chan \etal \cite{chan2019metafusion} have shown that meta classification can be used to improve network accuracy, an idea that so-far has not been achieved for object detection.
Previous studies have overlooked class-restricted meta classification performance, \eg, when restricting to safety-relevant instance classes.
Moreover, in order to base downstream applications on meta classification outputs, resulting confidences need to be statistically reliable, \ie, calibrated which has escaped previous research.

\section{Gradient-based epistemic uncertainty}
\label{sec: gradient uncertainty}
In instance-based recognition tasks, such as object detection or instance segmentation, the prediction consists of a list 
\begin{equation}
    \label{eq: od prediction}
    \hat{y} = (\hat{y}^1, \ldots, \hat{y}^{N_{\x}})
\end{equation}
of instances (\eg, bounding boxes).
The length of $\hat{y}$ usually depends on the corresponding input $\x$ and on hyperparameters (\eg, confidence / overlap thresholds).
Uncertainty information which is not generated directly from instance-wise data such as activation- or gradient-based information can \emph{at best yield statements about the entirety of $\hat{y}$} but not immediately about individual instances $\hat{y}^j$.
This issue is especially apparent for uncertainty generated from deep features which potentially all contribute to an instance $\hat{y}^j$.
Here, we introduce an approach to generate gradient-based uncertainty metrics for the instance-based setting.
To this end, we sketch how gradient uncertainty is generated for classification tasks.

Generically, given an input \(\x\), a classification network predicts a class distribution $\hat{y}(\x, \w) = (\hat{p}_1, \ldots, \hat{p}_C)$ of fixed length $C$ given a set of weights \(\w\).
During training, the latter is compared to the ground truth label \(y\) belonging to \(\x\) by means of some loss function \(\L(\cdot, \cdot)\), which is minimized by optimizing $\w$, \eg, by standard stochastic gradient descent.
The $\w$-step is proportional to the gradient \(g(\x, \w, y) := \nabla_{\!\w} \L(\hat{y}(\x, \w); y)\) which can also be regarded as a measure of \emph{learning stress} imposed upon $\w$.
Gradient uncertainty features are generated by substituting the non-accessible ground truth $y$ with the network's class prediction $\overline{y} := \argmax_c \{\hat{p}_c\}_{c = 1}^C$ and disregarding the dependence of the latter on $\w$.
In the following we will identify $\overline{y}$ with its one-hot encoding.
Scalar values are obtained by computing some magnitude of 
\begin{equation}
    \label{eq: gradient with replacement}
    g(\x, \w, \overline{y}) = \nabla_{\!\w} \L(\hat{y}(\x, \w), \overline{y}).
\end{equation}
To this end, in our experiments we employ the maps
\begin{equation}
    \label{eq: all maps for gradient metrics}
    \{\min(\cdot), \max(\cdot), \mean(\cdot), \std(\cdot), \norm{\cdot}_1, \norm{\cdot}_2\}.
\end{equation}
We discuss the latter choice in our supplementary material and first illuminate a couple of points about the expression in \cref{eq: gradient with replacement}.

\textbf{Intuition and discussion of \eqref{eq: gradient with replacement}.}
First of all, \cref{eq: gradient with replacement} can be regarded as the \emph{self-learning gradient} of the network.
It, therefore, expresses the learning stress on $\w$ under the condition that the class prediction $\overline{y}$ were given as the ground truth label.
The collapse of the (\eg, softmax) prediction $\hat{y}$ to $\overline{y}$ implies that \eqref{eq: gradient with replacement} does not generally vanish in the classification setting.
However, this consideration poses a problem for (bounding box) regression which we will address in the next paragraph.
We also note that it is possible to generate fine-grained metrics by restricting $\w$ in \eqref{eq: gradient with replacement} to sub-sets of weights $\w_\ell$, \eg, individual layers, convolutional filters or singular weights (computing partial gradients of $\L$).

Using \cref{eq: gradient with replacement} as a measure of uncertainty may be understood by regarding true and false predictions.
A well-performing network which has $\overline{y}$ already close to the true label $y$ tends to experience little stress when trained on $(\x, y)$ with the usual learning gradient.
This reflects \emph{confidence} in the prediction $\overline{y}$ and the difference between \cref{eq: gradient with replacement} and the true gradient is then small.
In the case of false predictions $\overline{y} \neq y$, the true learning gradient enforces large adjustments in $\w$.
The self-learning gradient \eqref{eq: gradient with replacement} behaves differently in that it is \emph{large for non-peaked/uncertain} (high entropy) predictions $\hat{y}$ and small for highly peaked distributions.

\textbf{Extension to object detectors.}

We first clarify the aforementioned complications in generating uncertainty information for object detection.
Generally, the prediction \eqref{eq: od prediction} is the filtering result of a larger, often fixed number \(\hat{N}_{\mathrm{out}}\) of output bounding boxes $\widetilde{y}(\x, \w)$ (refer to \cref{app: object detection} for details).
Given a ground truth list $y$ of bounding boxes, the loss function usually has the form
\begin{equation}
    \L = \L(\widetilde{y}(\x, \w); y),
\end{equation}
such that all \(\hat{N}_{\mathrm{out}}\) output bounding boxes potentially contribute to \(g(\x, \w, y)\).
Again, when filtering $\widetilde{y}$ to a smaller number of predicted boxes $\hat{y}$ and converting them to ground truth format $\overline{y}$, we can compute the self-learning gradient $g(\x, \w, \overline{y})$.
This quantity, however, does not refer to any individual prediction $\hat{y}^j$, but rather to all boxes in $\overline{y}$ simultaneously.
We take \emph{two steps to obtain meaningful gradient information} for one particular box $\hat{y}^j$ from this approach.

\emph{Firstly}, we restrict the ground truth slot to only contain the length-one list $\overline{y}^j$, regarding it as the hypothetical label.
This alone is insufficient since other, correctly predicted instances in $\widetilde{y}(\x,\w)$ would lead to a penalization and ``overcorrecting'' gradient $g(\x, \w, \overline{y}^j)$, given $\overline{y}^j$ as label.
This gradient's optimization goal is, figuratively speaking, to forget to predict everything but $\hat{y}^j$ when presented with $\x$.
Note that we cannot simply compute $\nabla_{\!\w} \L(\hat{y}^j(\x, \w); \overline{y}^j)$ since regression losses, such as for bounding box regression, are frequently norm-based (\eg, $L^p$-losses, see \cref{app: loss functions}) such that the respective loss and gradient would both vanish.
Therefore, we \emph{secondly} mask $\widetilde{y}$ such that the result is likely to only contain output boxes meaning to predict the same instance as $\overline{y}^j$.
Our conditions for this mask are \emph{sufficient score}, \emph{sufficient overlap} with $\overline{y}^j$ and \emph{same indicated class} as $\overline{y}^j$.
We call the subset of $\widetilde{y}$ that satisfies these conditions \emph{candidate boxes} for $\overline{y}^j$, denoted $\mathrm{cand}[\overline{y}^j]$ (for details, refer to \cref{app: object detection}).
We, thus, propose the candidate-restriced self-learning gradient
\begin{equation}
    \label{eq: gradient with candidates}
    g^{\mathrm{cand}}(\x, \w, \hat{y}^j) := \nabla_{\!\w} \L\left(\mathrm{cand}[\hat{y}^j](\x, \w), \overline{y}^j\right)\end{equation}
of $\hat{y}^j$ for computing instance-wise uncertainty.
This approach is in in line with the motivation for the classification setting and extends it when computing \eqref{eq: gradient with candidates} for individual contributions to the multi-criterial loss function in object detection (see \cref{app: loss functions}).

\textbf{Computational complexity.}
Sampling-based epistemic uncertainty quantification methods such as MC dropout and deep ensembles tend to generate a significant computational overhead as several forward passes are required.
Here, we provide a theoretical result on the count of floating point operations (FLOP) of gradient uncertainty metrics which is supported with a proof and additional detail in \cref{app: complexity}.
In our experiments, we use the gradients computed over the last two layers of each network architecture (perhaps of different architectural branches, as well; details in \cref{app: implementation details}).
For layer $t$, we assume stride-1, $(2 s_t + 1) \times (2 s_t + 1)$-convolutional layers acting on features maps of spatial size $w_t \times h_t$.
These assumptions hold for all architectures in our experiments.
We denote the number of input channels by $k_{t - 1}$ and of output channels by $k_t$.
\begin{thm}
    \label{thm: complexity}
    The number of FLOP required to compute the last layer ($t = T$) gradient in \cref{eq: gradient with candidates} 
    is \(\mathcal{O}(k_T h w + k_T k_{T - 1}(2s_T+1)^4)\).
    Similarly, for earlier layers $t$, 
    we have \(\mathcal{O}(k_{t + 1} k_{t} + k_{t} k_{t - 1})\), provided that we have previously computed the gradient for the consecutive layer \(t + 1\).
    
    Performing variational inference only on the last layer requires \(\mathcal{O}(k_T k_{T - 1} h w)\) FLOP per sample.
\end{thm}
\Cref{thm: complexity} provides that even for MC dropout before the last layer, or the use of efficient deep sub-ensembles \cite{valdenegrodeep} sharing the entire architecture but the last layer, gradient metrics require fewer or at worst similar FLOP counts.
Earlier sampling, especially entire deep ensembles, have even higher FLOP counts than these variants.
Note, however, that computing gradient metrics have somewhat larger computational latency since the full forward pass needs to be computed before the loss gradient can be computed.
Moreover, while sampling strategies can in principle be implemented to run all sample forward passes in parallel, the computation of gradients can in principle run in parallel for predicted boxes per image (see \cref{app: loss functions}).

\section{Meta classification and meta regression}
\label{sec: methods}
\newcommand{\M}{\mathcal{D}}
\begin{figure*}
    \centering
    \resizebox{\linewidth}{!}{
    \input{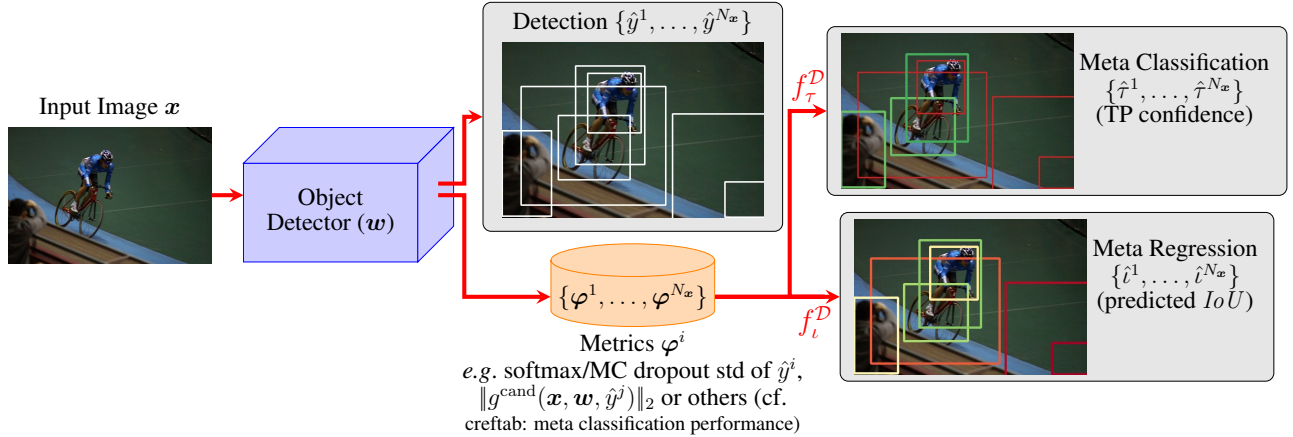}
    }
    \caption{
    Meta classification and meta regression pipeline:
    An uncertainty feature vector $\bm{\varphi}^j$ is assigned to each detected box $\hat{y}^j$.
    During training, we fit $f_\tau^{\M}$ and $f_\iota^{\M}$ to map $\bm{\varphi}^j$ to $\tau^j$ (TP/FP) and max.\ $\iou$ $\iota^j$ of $\hat{y}^j$, resp.
    At inference, $f_\tau^{\M}$ and $f_\iota^{\M}$ yield confidence and $\iou$ estimates $\hat{\tau}^j$ and $\hat{\iota}^j$ for $\hat{y}^j$ based on $\bm{\varphi}^j$.
    }
    \label{fig: meta class + meta reg pipeline}
\end{figure*}

We evaluate the efficacy of gradient metrics in terms of meta classification and meta regression.
These two approaches allow for the \emph{aggregation of potentially large feature vectors to obtain uncertainty estimates} for a respective prediction (\eg, a bounding box).
The aim of meta classification is to detect FP predictions by generating confidence estimates while meta regression directly estimates the prediction quality (\eg, $\iou$).
This, in turn, allows for the unified comparison of different uncertainty quantification methods and combinations thereof by regarding meta classifiers and meta regression models based on different features.
Moreover, we are able to investigate the degree of mutual redundancy of different sources of uncertainty. 
In the following, we summarize this method for bounding box detection and illustrate the scheme in \cref{fig: meta class + meta reg pipeline}.

We regard an object detector generating a list of $N_{\x}$ detections along with a vector $\bm\varphi^j$ for each predicted bounding box $\hat{y}^j$.
This vector $\bm\varphi^j \in \R^n$ of $n$ \enquote{metrics} may contain gradient metrics, but also, \eg, bounding box features, MC dropout or deep ensemble features or combinations thereof (\eg, by concatenation of dropout and ensemble feature vectors).
On training data $\M$, we compute boxes $\hat{y}$ and corresponding metrics $\varphi = (\bm\varphi^1, \ldots, \bm \varphi^{N_{\x}})$.
We evaluate each predicted instance $\hat{y}^j$ corresponding to the metrics $\varphi^j$ in terms of their maximal $\iou$, denoted $\iota^j \in[0, 1]$ with the respective ground truth and determine FP/TP labels $\tau^j \in \{0, 1\}$ (see \Cref{app: object detection}).
A \emph{meta classifier} is a lightweight classification model $f_\tau : \R^n \to (0, 1)$ giving probabilities for the classification of $\bm\varphi^j$ (vicariously for the uncertainty of $\hat{y}^j$) as TP which we fit on $\M$.
Similarly, a \emph{meta regression} model $f_\iota : \R^n \to \R$ is fit to the maximum $\iou$ $\iota^j$ of $\hat{y}^j$ with the ground truth of $\x$.
The models $f_\tau^\M$ and $f_\iota^\M$ can be regarded as post-processing modules which generate confidence measures given an input to an object detector leading to features $\varphi^j$.
At inference time, we then obtain box-wise classification probabilities $\hat{\tau}^k = f_\tau^\M(\varphi^k)$ and $\iou$ predictions $\hat{\iota}^k = f_\iota^\M(\varphi^k)$.
We then determine the predictive power of $f_\tau^\M$ and $f_\iota^\M$ in terms of  their area under receiver operating characteristic ($\auroc$) or average precision ($\ap$) scores and the coefficient of determination ($R^2$) scores, respectively.

\textbf{MetaFusion (object detection post-processing).}
As a direct application of uncertainty metrics, we investigate an approach inspired by \cite{chan2019metafusion}.
We \emph{implement meta classification into the object detection pipeline} by assigning each output box in $\widetilde{y}$ its meta classification probability as prediction confidence as shown in \cref{fig: reliability classifiers}.

State-of-the-art object detectors use score thresholding in addition to NMS which we compare with confidence filtering based on meta classification.
Since for most competitive uncertainty baselines in our experiments, computation for the entire pre-filtering network output $\widetilde{y}$ is expensive, we implement a small score threshold which still allows for a large amount of predicted boxes (of $\sim 150$ bounding boxes per image).

This way, well-performing meta classifiers (which accurately detect FPs) together with an increase in detection sensitivity offer a way to \enquote{trade} uncertainty information for detection performance.

Note, that in most object detection pipelines, score thresholding is carried out before NMS.
We choose to interchange them here as they commute for the baseline approach.
The resulting predictions are compared for a range of confidence thresholds in terms of mean Average Precision (\(\map\) \cite{Everingham10}).

\section{Experiments}
\label{sec: experiments}
\begin{table}
    \centering
    \caption{Number of layers and losses utilized and resulting numbers of gradient metrics per box.
        Multiplication in \# layers denotes parallel output strands of the resp.\ DNN (no additional gradients).}
    \label{tab: number of layers losses and splits}
    \resizebox{0.9\linewidth}{!}{
    \begin{tabular}{l c c c}
        \toprule
        Architecture & \# layers & \# Losses & \# gradients \\ \midrule
        
        YOLOv3 & \(2 \times 3\) & $3$ & $6$ \\
        Faster R-CNN & \(2 \times 4\) & $4$ & $8$ \\
        
        RetinaNet & \(2 \times 2\) & $2$ & $4$ \\
        Cascade R-CNN & \(2 \times 8\) & $8$ & 16 \\
        \bottomrule
    \end{tabular}
    }
\end{table}

In this section, we report our numerical methods and experimental findings.
We investigate meta classification and meta regression on three object detection datasets, namely Pascal VOC \cite{Everingham10}, MS COCO \cite{lin2014microsoft} and KITTI \cite{geiger2015kitti}.
The splits for training and evaluation we used are given in \cref{app: implementation details}.
We investigate for gradient-based meta classification and meta regression for only 2-norm scalars, denoted $\mathrm{GS}_{\norm{\cdot}_2}$ (refer to \cref{sec: gradient uncertainty}), as well as the larger model for all maps listed in \cref{eq: all maps for gradient metrics}, denoted $\mathrm{GS}_{\mathrm{full}}$.
Gradient metrics are always computed for the last two network layers of each architectural branch and for each contribution to the loss function $\L$ separately, \ie, for classification, bounding box regression and, if applicable, objectness score.
We list the resulting counts and number of gradients per investigated architecture in \cref{tab: number of layers losses and splits}.
As meta classifiers and meta regressors, we use gradient boosting models which have been shown\cite{vasudevan2019towards,schubert2020metadetect,maag2020improving} to perform well as such.
For implementation details, we refer the reader to \cref{app: implementation details}.
Whenever we indicate means and standard deviations, we obtained those by 10-fold image-wise cross validation (cv) for the training split $\M$ of the meta classifier / meta regression model.
Evaluation is done on the complement of $\M$.

\begin{table*}
    \centering
    \caption{Meta classification performance in terms of \(\auroc\) and \(\ap\) per confidence model over 10-fold cv (\(\mathrm{mean} \pm \std\)).
    }
    \resizebox{0.9\textwidth}{!}{
    \begin{tabular}{l | c c  c c  c c}\toprule
                    \textbf{YOLOv3}    &   \multicolumn{2}{c}{Pascal VOC}      &   \multicolumn{2}{c}{COCO}                &   \multicolumn{2}{c}{KITTI}   \\\midrule
                        &   \(\auroc\)           &   \(\ap\)  &   \(\auroc\)               &   \(\ap\)  &   \(\auroc\)               &   \(\ap\)   \\\midrule
        \rowcolor{LightGray}
        Score           &   $90.68 \pm 0.06$ & $69.56 \pm 0.12$ &   $82.97 \pm 0.04$ & $62.31 \pm 0.05$ &   $96.53 \pm 0.05$     &   $96.87 \pm 0.03$   \\
        Entropy         & $91.30 \pm 0.02$ & $61.94 \pm 0.06$ & $76.52 \pm 0.02$ & $42.52 \pm 0.04$ & $94.79 \pm 0.06$ & $94.83 \pm 0.05$ \\
        \rowcolor{LightGray}
        Energy Score \cite{liu2020energy} & $92.59 \pm 0.02$ & $64.65 \pm 0.06$ & $75.39 \pm 0.02$ & $39.72 \pm 0.06$ & $95.66 \pm 0.02$ & $95.33 \pm 0.03$ \\
        Full Softmax & $93.81 \pm 0.06$ & $72.08 \pm 0.15$ & $82.91 \pm 0.06$ & $58.65 \pm 0.10$ & $97.07 \pm 0.03$ & $96.85 \pm 0.03$ \\
        \rowcolor{LightGray}
        MC Dropout \cite{srivastava2014dropout} (MC) &   \underline{$96.72 \pm 0.02$} & $78.15 \pm 0.09$ & $\mathbf{89.04 \pm 0.02}$ & $64.94 \pm 0.11$ &   $97.60 \pm 0.07$     &   $97.17 \pm 0.10$   \\
        Ensemble \cite{lakshminarayanan2016simple} (E) & $\mathbf{96.87 \pm 0.02}$ & $77.86 \pm 0.11$ & \underline{$88.97 \pm 0.02$} & $64.05 \pm 0.12$ & $97.63 \pm 0.04$ & $97.63 \pm 0.05$ \\
        \rowcolor{LightGray}
        MetaDetect \cite{schubert2020metadetect} (MD) &   $95.78 \pm 0.05$ & $\mathbf{78.64 \pm 0.08}$ &   $87.16 \pm 0.04$     &   \underline{$69.41 \pm 0.07$} &   $\mathbf{98.23 \pm 0.02}$     &   $\mathbf{98.06 \pm 0.02}$   \\
        Grad.\ Score\({}_{\norm{\cdot}_2}\) (GS\({}_{\norm{\cdot}_2}\); ours)        &   $94.76 \pm 0.03$ & $74.86 \pm 0.10$ &   $86.05 \pm 0.04$     &   $64.25 \pm 0.06$ &   $97.31 \pm 0.05$     &   $96.86 \pm 0.10$   \\
        \rowcolor{LightGray}
        Grad.\ Score\({}_{\mathrm{full}}\) (GS\({}_{\mathrm{full}}\); ours)  &   $95.80 \pm 0.04$ & \underline{$78.57 \pm 0.11$} &   $88.07 \pm 0.03$     &   $\mathbf{69.62 \pm 0.07}$ &   \underline{$98.04 \pm 0.03$}     &   \underline{$97.81 \pm 0.06$}   \\\midrule
        \rowcolor{LightCyan}
        MC+E+MD (ours) & $97.66 \pm 0.02$ & $85.13 \pm 0.12$ & $91.14 \pm 0.02$ & $73.82 \pm 0.05$ & $98.56 \pm 0.03$ & $98.45 \pm 0.03$ \\
        \rowcolor{LightCyan}
        GS\({}_{\mathrm{full}}\)+MC+E+MD (ours) & $\mathbf{97.95 \pm 0.02}$ & $\mathbf{86.69 \pm 0.09}$ & $\mathbf{91.65 \pm 0.03}$ & $\mathbf{74.88 \pm 0.07}$ & $\mathbf{98.74 \pm 0.02}$ & $\mathbf{98.62 \pm 0.01}$ \\
        \bottomrule
    \end{tabular}
    }
    \label{tab: meta classification performance}
\end{table*}
\begin{table}
    \centering
    \caption{Meta regression performance in terms of \(R^2\) per confidence model over 10-fold cv (\(\mathrm{mean} \pm \std\)).}
    \resizebox{\linewidth}{!}{
    \begin{tabular}{l | c c c }\toprule
                \textbf{YOLOv3}    &   Pascal VOC      &   COCO            &   KITTI  \\\midrule
    \rowcolor{LightGray}
    Score           &   $48.29 \pm 0.04$ &   $32.60 \pm 0.02$ &   $78.86 \pm 0.05$ \\
    Entropy        & $43.24 \pm 0.03$ & $21.10 \pm 0.04$ & $69.33 \pm 0.04$  \\
    \rowcolor{LightGray}
    Energy Score    & $47.18 \pm 0.03$ & $17.94 \pm 0.02$ & $71.53 \pm 0.10$  \\
    Full Softmax    & $53.86 \pm 0.11$ & $36.95 \pm 0.13$ & $78.92 \pm 0.11$  \\
    \rowcolor{LightGray}
    MC              &   \underline{$61.63 \pm 0.15$} &   $43.85 \pm 0.09$ &   $82.10 \pm 0.11$ \\
    
    E  & $61.48 \pm 0.07$ & $43.53 \pm 0.13$ & $84.18 \pm 0.12$ \\
    \rowcolor{LightGray}
    MD              &   $60.36 \pm 0.14$ &   \underline{$44.22 \pm 0.11$} &   $\mathbf{85.88 \pm 0.10}$ \\
    GS\({}_{\norm{\cdot}_2}\) (ours)    &   $58.05 \pm 0.13$ &   $38.77 \pm 0.04$ &   $81.21 \pm 0.05$ \\
    
    \rowcolor{LightGray}
    GS\({}_\mathrm{full}\) (ours)  &   $\mathbf{62.50 \pm 0.11}$ &   $\mathbf{44.90 \pm 0.09}$ &   \underline{$85.40 \pm 0.11$} \\ \midrule
    \rowcolor{LightCyan}
    MC+E+MD & $69.38 \pm 0.11$ & $54.07 \pm 0.08$ & $87.78 \pm 0.11$\\
    \rowcolor{LightCyan}
    GS\({}_\mathrm{full}\)+MC+E+MD  & $\mathbf{72.26 \pm 0.08}$ & $\mathbf{56.14 \pm 0.11}$ & $\mathbf{88.80 \pm 0.07}$  \\

    \bottomrule
    \end{tabular}
    }
    \label{tab: meta regression performance}
\end{table}
\textbf{Comparison with output-based uncertainty.}
We compare gradient-based uncertainty with various uncertainty baselines in terms of meta classification (\cref{tab: meta classification performance}) and meta regression (\cref{tab: meta regression performance}) for a YOLOv3 model with standard Darknet53 backbone \cite{redmon2018yolov3}.
As class probability baselines, we consider objectness score, softmax entropy, energy score \cite{liu2020energy} and the full softmax distribution per box.
Since the full softmax baseline fits a model directly to all class probabilities (as opposed to relying on hand-crafted functions), it can be considered an \emph{enveloping model} to both, entropy and energy score.
Moreover, we consider other output baselines in MC dropout (MC), deep ensembles (E) and MetaDetect (MD; details in \cref{app: implementation details}).
Since MetaDetect involves the entire network output of a bounding box, it leads to meta classifiers fitted on more variables than class probability baselines.
It is, thus, an enveloping model of the full softmax baseline and, therefore, all classification baselines.
The results in \cref{tab: meta classification performance} indicate that $\mathrm{GS}_{\mathrm{full}}$ is roughly in the same $\auroc$ range as sampling-based uncertainty methods, while being consistently among the two best methods in terms of $\ap$.
The smaller gradient-based model $\mathrm{GS}_{\norm{\cdot}_2}$ is consistently better than the full softmax baseline, by up to $3.14$ $\auroc$ percentage points (ppts) and up to $5.60$ $\ap$ ppts.

We also find that $\mathrm{GS}_{\mathrm{full}}$ tends to rank lower in terms of $\auroc$.
Note also, that MetaDetect is roughly on par with the sampling approaches MC and E throughout.
While the latter methods all aim at capturing epistemic uncertainty they constitute approximations and are, thus, not necessarily mutually redundant (cf.\\cref{app: results}).

In addition, we compare the largest sampling and output based model in MC+E+MD and add the gradient metrics $\mathrm{GS}_{\mathrm{full}}$ to find out about the degree of redundancy between the approximated epistemic uncertainty in MC+E+MD and our method.
We note significant boosts to the already well-performing model MC+E+MD across all metrics.
\Cref{tab: meta regression performance} suggests that gradient uncertainty is especially informative for meta regression with $\mathrm{GS}_{\mathrm{full}}$ being consistently among the best two models and achieving $R^2$ scores of up to $85.4$ on the KITTI dataset.
Adding gradient metrics to MC+E+MD always leads to a gain of more than one $R^2$ ppt indicating non-redundancy of gradient- and sampling-based metrics.

\begin{table*}[t]
    \centering
    \caption{Meta classification and meta regression performance in terms of \(\auroc\) and \(R^2\), respectively, for different object detection architectures.
    Results ($\mathrm{mean} \pm \std$) obtained from 10-fold cv as above.
    }
    \resizebox{0.8\textwidth}{!}{
    \begin{tabular}{l | c c  c c  c c}\toprule
                    &   \multicolumn{2}{c}{Pascal VOC}      &   \multicolumn{2}{c}{COCO}                &   \multicolumn{2}{c}{KITTI}   \\
        \midrule
                        &   \(\auroc\)           &   \(R^2\)  &   \(\auroc\)               &   \(R^2\)  &   \(\auroc\)               &   \(R^2\)   \\\midrule
        \textbf{Faster R-CNN} & \\\midrule
        
        Score           &   $89.77 \pm 0.05$ &   $39.94 \pm 0.02$ &   $83.82 \pm 0.03$     &   $40.50 \pm 0.01$ &   $96.53 \pm 0.05$     &   $72.29 \pm 0.02$   \\
        MD    &  $94.43 \pm 0.02$ &  $47.92 \pm 0.09$ &  $91.31 \pm 0.02$ &  $44.41 \pm 0.04$ &  $98.86 \pm 0.02$ &  $79.92 \pm 0.04$ \\
        
        GS\({}_{\mathrm{full}}\)    & $95.88 \pm 0.05$ &  $59.40 \pm 0.03$ &  $91.38 \pm 0.03$ &  $50.44 \pm 0.04$ &  $99.20 \pm 0.01$ &  $86.31 \pm 0.07$  \\
        \rowcolor{LightCyan}
        GS\({}_{\mathrm{full}}\) + MD &  $96.77 \pm 0.05$ &  $63.64 \pm 0.08$ &  $92.30 \pm 0.02$ &  $52.30 \pm 0.04$ &  $99.37 \pm 0.02$ &  $87.46 \pm 0.05$  \\
        \midrule
        \textbf{RetinaNet} & \\\midrule
        
        Score           &   $87.53 \pm 0.03$ &   $40.43 \pm 0.01$ & $84.95 \pm 0.02$   &   $39.88 \pm 0.02$ &   $95.91 \pm 0.02$     &   $73.44 \pm 0.02$   \\
        MD    &  $89.57 \pm 0.04$ &  $50.27 \pm 0.10$ &  $85.09 \pm 0.01$ &  $42.45 \pm 0.12$ &  $96.19 \pm 0.02$ &  $77.53 \pm 0.08$  \\
        
        GS\({}_{\mathrm{full}}\)    &  $91.58 \pm 0.04$ &  $57.23 \pm 0.07$ &  $85.59 \pm 0.02$ &  $47.74 \pm 0.06$ &  $97.26 \pm 0.03$ &  $84.47 \pm 0.04$  \\
        \rowcolor{LightCyan}
        GS\({}_{\mathrm{full}}\) + MD &  $92.99 \pm 0.03$ &  $64.32 \pm 0.07$ &  $87.15 \pm 0.05$ &  $51.07 \pm 0.09$ &  $97.61 \pm 0.02$ &  $85.73 \pm 0.09$   \\
        \midrule
        \textbf{Cascade R-CNN} & \\\midrule
        
        Score           &   $95.70 \pm 0.04$ &   $57.90 \pm 0.09$ &   $94.11 \pm 0.01$     &   $56.31 \pm 0.01$ &   $98.67 \pm 0.02$     &   $83.31 \pm 0.03$   \\
        MD    &   $96.32 \pm 0.05$ &   $63.62 \pm 0.12$ &   $94.10 \pm 0.02$     &   $58.74 \pm 0.08$ &   $99.18 \pm 0.01$     &   $86.22 \pm 0.08$ \\
        
        GS\({}_{\mathrm{full}}\)    & $96.66 \pm 0.05$ & $63.94 \pm 0.13$ & $93.97 \pm 0.01$ & $57.80 \pm 0.08$ & $99.34 \pm 0.01$ & $87.39 \pm 0.08$  \\
        \rowcolor{LightCyan}
        GS\({}_{\mathrm{full}}\) + MD & $97.24 \pm 0.05$ & $69.78 \pm 0.13$ & $94.78 \pm 0.02$ & $62.13 \pm 0.06$ & $99.48 \pm 0.01$ & $89.59 \pm 0.04$  \\
        \bottomrule
    \end{tabular}
    }
    \label{tab: meta performance networks}
\end{table*}
\textbf{Object detection architectures.}
We investigate the applicability and viability of gradient uncertainty for a variety of different architectures.
In addition to the YOLOv3 model, we investigate two more standard object detectors in Faster R-CNN \cite{ren2015faster} and RetinaNet \cite{lin2017focal} both with a ResNet50 backbone \cite{he2016deep}.
Moreover, we investigate a stronger object detector in Cascade R-CNN \cite{cai2018cascade} with a large ResNeSt200 \cite{zhang2020resnest} backbone which at the time of writing was ranked among the top 10 on the \href{https://cocodataset.org/\#detection-leaderboard}{official COCO Detection Leaderboard}.
With a COCO detection $\mathit{AP}$ of $49.03$, this is in the state-of-the-art range for pure, non-hybrid-task object detectors.
In \cref{tab: meta performance networks}, we list meta classification $\auroc$ and meta regression $R^2$ for the score, MetaDetect (representing output-based methods), $\mathrm{GS}_{\mathrm{full}}$ and the combined model $\mathrm{GS}_{\mathrm{full}}$+MD.
We see $\mathrm{GS}_{\mathrm{full}}$ again being on par with MD, in the majority of cases even surpassing it by up to $2.01$ $\auroc$ ppts and up to $11.52$ $R^2$ ppts.
When added to MD, we find again boosts in both performance metrics, especially in $R^2$.
On the COCO dataset, the high performance model Cascade R-CNN delivers a remarkably strong Score baseline completely redundant with MD and surpassing $\mathrm{GS}_{\mathrm{full}}$ on its own.
However, here we also find an improvement of $0.68$ ppts by adding gradient information.

\textbf{Calibration.}
\begin{figure}
    \centering
    \includegraphics[width=\linewidth]{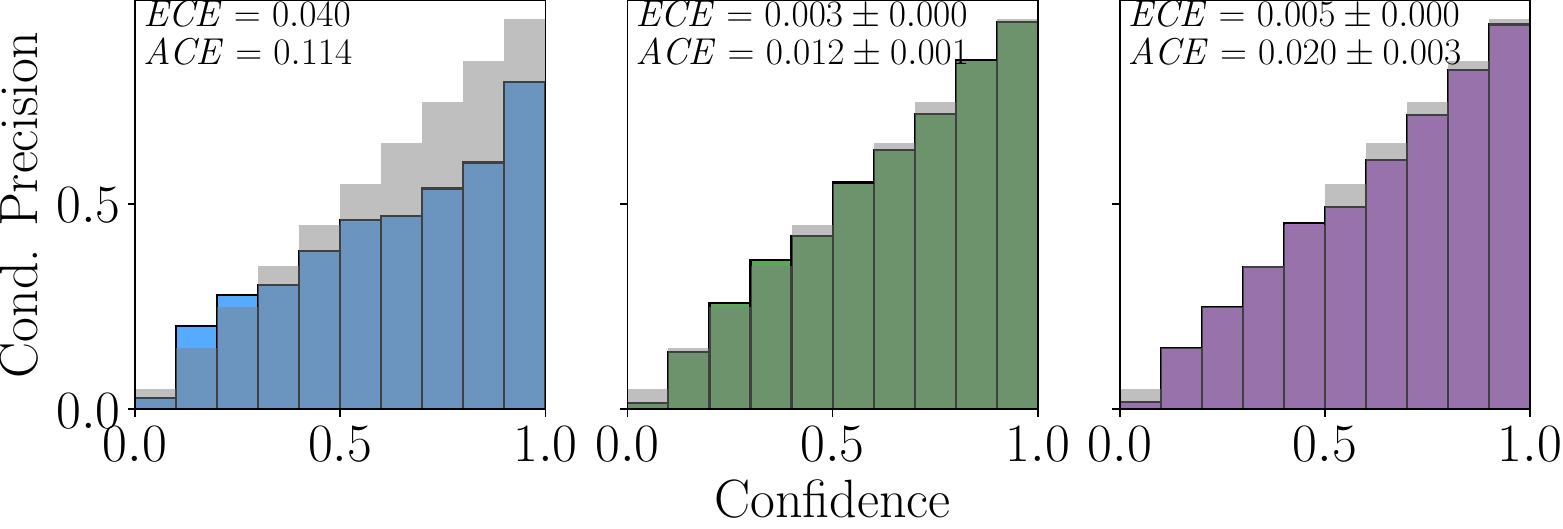}
    \caption{Reliability plots of the Score (left) and meta classifiers for MD (center) and $\mathrm{GS}_{\mathrm{full}}$ (right) on the Pascal VOC dataset  for YOLOv3 with calibration errors ($\mathrm{mean} \pm \std$).
    }
    \label{fig: reliability example}
\end{figure}
We evaluate the meta classifier confidences obtained above in terms of their calibration errors when divided into 10 confidence bins.
Exemplary reliability plots are shown in \cref{fig: reliability example} for the Score, MD and $\mathrm{GS}_{\mathrm{full}}$ together with corresponding expected ($\ECE$\cite{naeini2015obtaining_bayesian_binning}) and average ($\ACE$\cite{neumann2018relaxed}) calibration errors.
The Score is clearly over-confident in the upper confidence range and both meta classifiers are well-calibrated.
Both calibration errors of the latter are about one order of magnitude smaller than those of the score.
We show the full table of calibration errors including maximum calibration error in \cref{app: results}.

\begin{figure}
    \centering
    \includegraphics[width=\linewidth]{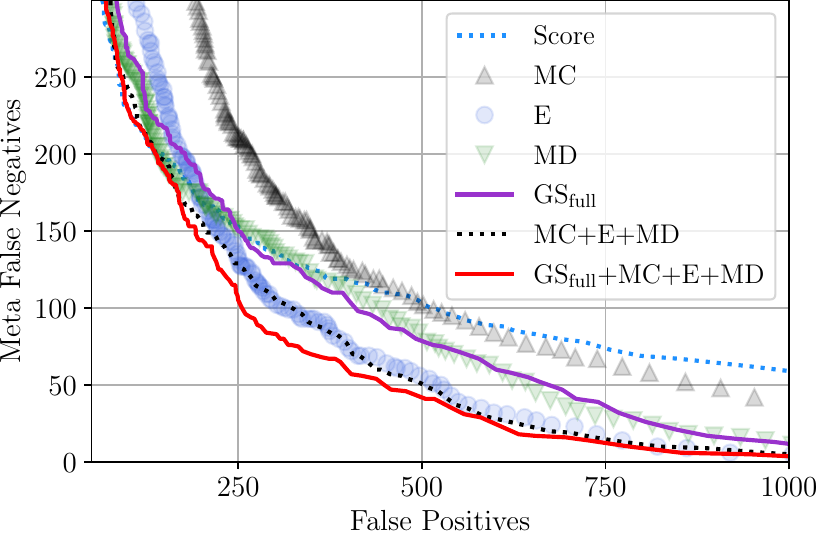}
    \hfill
    \caption{Meta classification for the class \enquote{Pedestrian}. 
    Curves obtained by sweeping the threshold on score / meta classification probability.
    Note the FP gaps for $\leq 100$ FNs.
    }
    \label{fig: pareto pedestrian plot}
\end{figure}

\begin{figure}
    \centering
    \includegraphics[width=\linewidth]{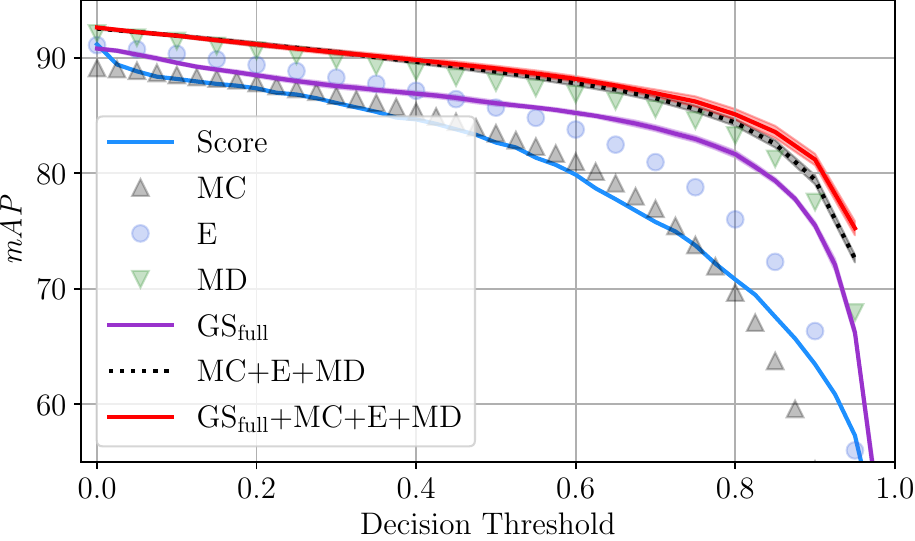}
    \caption{Score baseline and MetaFusion \(\map\). Error bands we draw around meta classifiers indicate cv-$\std$.}
    \label{fig: meta fusion map plot}
\end{figure}

\textbf{Pedestrian detection.}
The statistical improvement seen in meta classification performance may not hold for non-majority classes within a dataset which are regularly safety-relevant.
We investigate meta classification of the ``Pedestrian'' class in KITTI and explicitly study the FP/FN trade-off.
This can be accomplished by sweeping the confidence threshold between $0$ and $1$ and counting the resulting FPs and FNs.
We choose increments of $10^{-2}$ for meta classifiers and $10^{-4}$ for the scores as to not interpolate too roughly in the range of very small score values where a significant number of predictions cluster.
The resulting curves are depicted in \cref{fig: pareto pedestrian plot}.
For applications in safety-critical environments, not all errors need to be equally important.
We may, for example, demand a good trade-off at a given FN count which is usually desired to be especially small.
Our present evaluation split contains a total of $1152$ pedestrian instances.
Assume that we allowed for a detector to miss around $100$ pedestrians ($\sim 10\%$), we see a reduction in FPs for some meta classifiers.  
MD and $\mathrm{GS}_{\mathrm{full}}$ are very roughly on par, leading to a reduction of close to $100$ FPs.
The ensemble E turns out to be about as effective as the entire output-based model MC+E+MD, only falling behind above $150$ FNs.
This indicates a certain degree of redundancy between output-based methods.
Adding $\mathrm{GS}_{\mathrm{full}}$ to MC+E+MD, however, reduces the number of FPs again by about $100$ leading to an FP difference of about $250$ as compared to the Score baseline.
Observing the trend, the improvements become even more effective for smaller numbers of FNs (small thresholds) but diminish for larger numbers of above $200$ FNs.

\textbf{MetaFusion.}

In regarding \cref{fig: meta class + meta reg pipeline}, meta classifiers naturally fit as post processing modules on top of object detection pipelines.
Doing so does not generate new bounding boxes, but modifies the confidence ranking as shown in \cref{fig: reliability classifiers} and may also lead to calibrated confidences.
Therefore, the score baseline and meta classifiers are not comparable for fixed decision thresholds.
We obtain a comparison of the resulting object detection performance by sweeping the decision threshold with a step size of $0.05$ (resp.\ $0.025$ for Score).
The $\map$ curves are shown in \cref{fig: meta fusion map plot}.
We draw error bands showing cv-$\std$ for $\mathrm{GS}_{\mathrm{full}}$, MC+E+MD and $\mathrm{GS}_{\mathrm{full}}$+MC+E+MD.
Meta classification-based decision rules are either on par (MC) with the score threshold or consistently allow for an $\map$ improvement of at least $1$ to $2$ $\map$ ppts.
In particular, MD performs well, gaining around 2 ppts in the maximum $\map$.
When comparing the addition of $\mathrm{GS}_{\mathrm{full}}$ to MC+E+MD, we still find slim improvements for thresholds $\geq 0.75$.
The score curve shows a kink at a threshold of $0.05$ and ends at the same maximum $\map$ as $\mathrm{GS}_{\mathrm{full}}$ while the confidence ranking is clearly improved for MC+E+MD and $\mathrm{GS}_{\mathrm{full}}$+MC+E+MD.
Note that meta classification based on $\mathrm{GS}_{\mathrm{full}}$ is less sensitive to the choice of threshold than the score in the medium range.
At a threshold of $0.3$ we have an $\map$ gap of about $1.4$ ppts which widens to $5.2$ ppts at $0.6$.

\section{Conclusion}
\label{sec: conclusion and outlook}
Applications of modern DNNs in safety-critical environments demand high performance on the one hand, but also reliable confidence estimation indicating where a model is not competent.

Novel uncertainty quantification methods tend to be developed in the simplified classification setting, the transfer of which to instance-based recognition tasks entails conceptual complications.
We have proposed and investigated a way of implementing gradient-based uncertainty quantification for deep object detection which complements output-based methods well and is on par with established epistemic uncertainty metrics.
Experiments involving a number of different architectures suggest that our method can be applied to significant benefit across architectures, even for high performance state-of-the-art models.
We showed that meta classification performance carries over to object detection performance when employed as a post-processing module and that meta classification naturally leads to well-calibrated gradient confidences which improves probabilistic reliability.
Our trade-off study indicates that gradient uncertainty reduces the FP/FN ratio for non-majority classes, especially when paired with output-based methods which, however, considerably increases computational overhead.

While our experiments indicate the viability of gradient-based uncertainty for deep object detection empirically, we believe that a well-founded theoretical justification would further greatly benefit and advance the research area.
Also, a comparison of gradient metrics in terms of OoD (or ``open set condition'') detections would be of great service and in line with previous work on gradient uncertainty \cite{oberdiek2018classification,vasudevan2019towards,huang2021importance}.
However, the very definition of OoD in the instance-based setting is still subject of contemporary research itself\cite{miller2018dropout,dhamija2020overlooked,joseph2021towards} and lacks a widely established definition.

\Cref{eq: gradient with candidates} can in principle be augmented to fit any DNN inferring and learning on an instance-based logic such as 3D bounding box detection or instance segmentation.
Further applications of our method may include uncertainty-based querying in active learning or the probabilistic detection of data annotation errors.
We hope that this work will inspire future progress in uncertainty quantification, probabilistic object detection and related areas.

\paragraph{Foreseeable impact.}
While we see great benefits in increasing the safety of human lives through improved confidence estimation, we also note that the usage of machine learning models in confidence calibration based on gradient information adds another failure mode, which, if applied in automated driving, could lead to fatal consequences, if false predictions of confidence occur.
It is, therefore, necessary to be aware of the low technology readiness level of the method introduced here.
Despite the evidence provided that our method can significantly reduce the number of false negative pedestrians of an object detector for certain datasets, application of our method in a safety-critical context requires extensive testing to demonstrate robustness, \eg, with respect to domain shifts, prior to any integration into applications.

\noindent \textbf{Technical Limitations.}
We have argued that while sampling-based uncertainty quantification methods can be run in parallel across models, gradients can be computed in parallel across predicted bounding boxes.

Nevertheless, we emphasize that the usage of gradient uncertainty metrics introduces a factor of computational latency in that metrics can be computed only after the prediction.
Sampling-based methods do not suffer from this mechanism which should be considered, \eg for real-time applications.

\paragraph{Acknowledgement.}
The research leading to these results is funded by the German Federal Ministry for Economic Affairs and Energy within the project ``Methoden und Maßnahmen zur Absicherung von KI basierten Wahrnehmungsfunktionen für das automatisierte Fahren (KI-Absicherung)''. The authors would like to thank the consortium for the successful cooperation. Furthermore, we gratefully acknowledge financial support by the state Ministry of Economy, Innovation and Energy of Northrhine Westphalia (MWIDE) and the European Fund for Regional Development via the FIS.NRW project BIT-KI, grant no. EFRE-0400216. The authors gratefully acknowledge the Gauss Centre for Supercomputing e.V. (\href{www.gauss-centre.eu}{www.gauss-centre.eu}) for funding this project by providing computing time through the John von Neumann Institute for Computing (NIC) on the GCS Supercomputer JUWELS at Jülich Supercomputing Centre (JSC).

{\small
\bibliographystyle{plain}
\bibliography{biblio}
}

\clearpage

\appendix
\section{Object Detection}
\label{app: object detection}
\subsection{Notation.}
\label{app: OD notation}
We regard the task of 2D bounding box detection on camera images.
Here, the detection
\begin{equation}
    \hat{y}(\x, \w) = (\hat{y}^1(\x, \w), \ldots, \hat{y}^{{N}_{\x}}(\x, \w))
    \in \R^{N_{\x} \times (4 + 1 + C)}
\end{equation}
on an input image \(\x\), depending on model weights \(\w\), consists of a number \(N_{\x} \in \N\) (dependent on the input \(\x\)) of instances \(\hat{y}^j\).
We give a short account of its constituents.
Each instance
\begin{equation}
    \hat{y}^j = (\hat\xi^j, \hat{s}^j, \hat{p}^j) \in \R^{4 + 1 + C}
\end{equation} consists of localizations \(\hat{\xi}^j = (\hat{\mathsf{x}}^j, \hat{\mathsf{y}}^j, \hat{\mathsf{w}}^j, \hat{\mathsf{h}}^j)\) encoded e.g.\ as center coordinates\(\hat{\mathsf{x}}\), \(\hat{\mathsf{y}}\) together with width \(\hat{\mathsf{w}}\) and height \(\hat{\mathsf{h}}\).
Moreover, a list of \(N_{\x}\) integers \(\hat \kappa \in \{1, \ldots, C\}\) represents the predicted categories for the object found in the respective boxes out of a pre-determined fixed list of \(C \in \N\) possible categories.
Usually, \(\hat \kappa\)  is obtained as the \(\argmax\) of a learnt probability distribution \(\hat p = (\hat p_1, \ldots, \hat p_C) \in (0, 1)^C\) over all \(C\) categories.
Finally, \(N_{\x}\) scores \(\hat s \in (0, 1)\) indicate the probability of each box being correct.

The predicted \(N_{\x}\) boxes are obtained by different filtering mechanisms as a subset of a fixed number \(N_\mathrm{out}\) (usually about \(10^5\) to \(10^6\)) of output boxes 
\begin{equation}
    \widetilde{y}(\x, \w) = (\widetilde{y}^1, \ldots, \widetilde{y}^{N_\mathrm{out}}).
\end{equation}
The latter are the regression and classification result of pre-determined \enquote{prior} or \enquote{anchor boxes}.
Predicted box localization \(\hat{\xi}\) is usually learned as offsets and width- and height scaling of fixed anchor boxes \cite{redmon2018yolov3, lin2017focal} or region proposals \cite{ren2015faster} (see \cref{app: loss functions}).
The most prominent examples (and the ones employed in all architectures we investigate) of filtering mechanisms are \emph{score thresholding} and \emph{Non-Maximum Suppression}.
By score thresholding we mean only allowing boxes which have \(\hat s \geq \varepsilon_s\) for some fixed threshold \(\varepsilon_s \geq 0\).

\subsection{Non-Maximum Suppression (NMS).}
\label{app: nms}
NMS is an algorithm allowing for different output boxes that have the same class and significant mutual overlap (meaning they are likely to indicate the same visible instance in \(\x\)) to be reduced to only one box.
Overlap is usually quantified as \emph{intersection over union} (\(\iou\)).
For two bounding boxes \(A\) and \(B\), their intersection over union is 
\begin{equation}
    \iou(A, B) = \frac{|A \,\cap\, B|}{|A \,\cup\, B|},
\end{equation}
\ie, the ratio of the area of intersection of two boxes and the joint area of those two boxes, where $0$ means no overlap and $1$ means the boxes have identical location and size.
Maximal mutual \(\iou\) between a predicted box \(\hat{y}^j\) and ground truth boxes \(y\) is also used to quantify the quality of the prediction of a given instance.

We call an output instance \(\hat{y}^i\) \enquote{candidate box} for another box \(\hat{y}^j\) if it fulfills the following requirements:
\begin{enumerate}
    \item score (\(\hat{s}^i \geq \varepsilon_s\) above a chosen, fixed threshold \(\varepsilon_s\))
    \item identical class \(\hat{\kappa}^i = \hat{\kappa}^j\)
    \item large mutual overlap \(\iou(\hat{y}^i, \hat{y}^j) \geq \varepsilon_{\iou}\) for some fixed threshold \(\varepsilon_{\iou} \geq 0\) (a widely accepted choice which we adopt is \(\varepsilon_{\iou} = 0.5\)).
\end{enumerate}
We denote the set of output candidate boxes for \(\hat{y}^j\) by \(\mathrm{cand}[\hat{y}^j]\).
Note, that we can also determine candidate boxes for an output box $\widetilde{y}^j$.
In NMS, all output boxes are sorted by their score in descending order.
Then, the box with the best score is selected as a prediction and all candidates for that box are deleted from the ranked list.
This is done until there are no boxes with \(\hat{s} \geq \varepsilon_s\) left.
Thereby selected boxes form the \(N_{\x}\) predictions.

\subsection{Training of object detectors.}
\label{app: OD training}
The \enquote{ground truth} or  \enquote{label} data \(y\) from which an object detector learns must contain localization information \(\xi^j\) for each of \(j = 1, \ldots, N_{\x}\) annotated instances on each data point \(\x\), as well as the associated \(N_{\x}\) category indices \(\kappa^j\).
Note that we denote labels by the same symbol as the corresponding predicted quantity and omit the hat (\(\hat{\,\cdot\,}\)).

Generically, deep object detectors are trained by stochastic gradient descent or some variant such as AdaGrad \cite{duchi2011adaptive}, or Adam \cite{kingma2014adam} by minimizing an empirical loss
\begin{equation}
    \label{eq: contribution split of loss}
    \L = \L_\xi + \L_s + \L_p \, .
\end{equation}
For all object detection frameworks which we consider here (and most architectures, in general) the loss function \(\L\) splits up additively into parts punishing localization inaccuracies (\(\widetilde\xi\)), score (\(\widetilde s\)) assignment to boxes (assigning large loss to high score for incorrect boxes and low score for correct boxes) and incorrect class probability distribution (\(\widetilde p\)), respectively.
We explicitly give formulas for all utilized loss functions in \cref{app: loss functions}.
The trainable weights \(\w\) of the model are updated in standard gradient descent optimization by
\begin{equation}
    \label{eq: gradient descent update}
    \w \leftarrow \w - \eta \nabla_{\!\w} \L(\widetilde{y}(\x, \w), y)
\end{equation}
where \(\eta\) is a learning rate factor.
We denote by \(g(\x, \w, y) := \nabla_{\!\w} \L(\widetilde{y}(\x, \w); y)\) the learning gradient on the data point \((\x, y)\).

\begin{figure*}
    \centering
    \includegraphics[width=\textwidth]{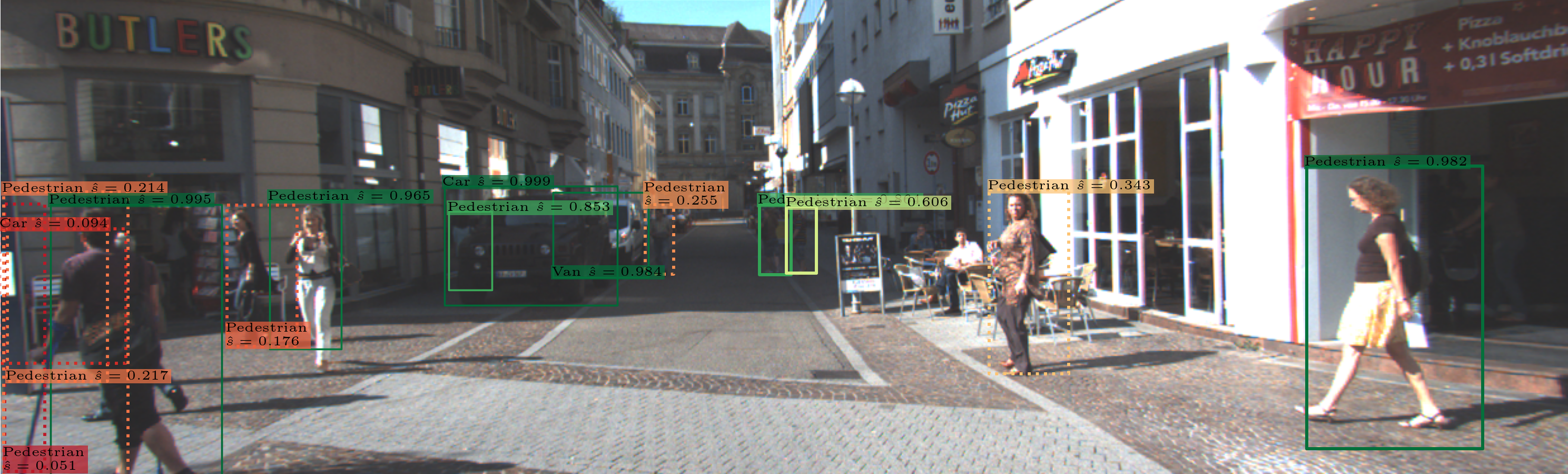}
    \includegraphics[width=\textwidth]{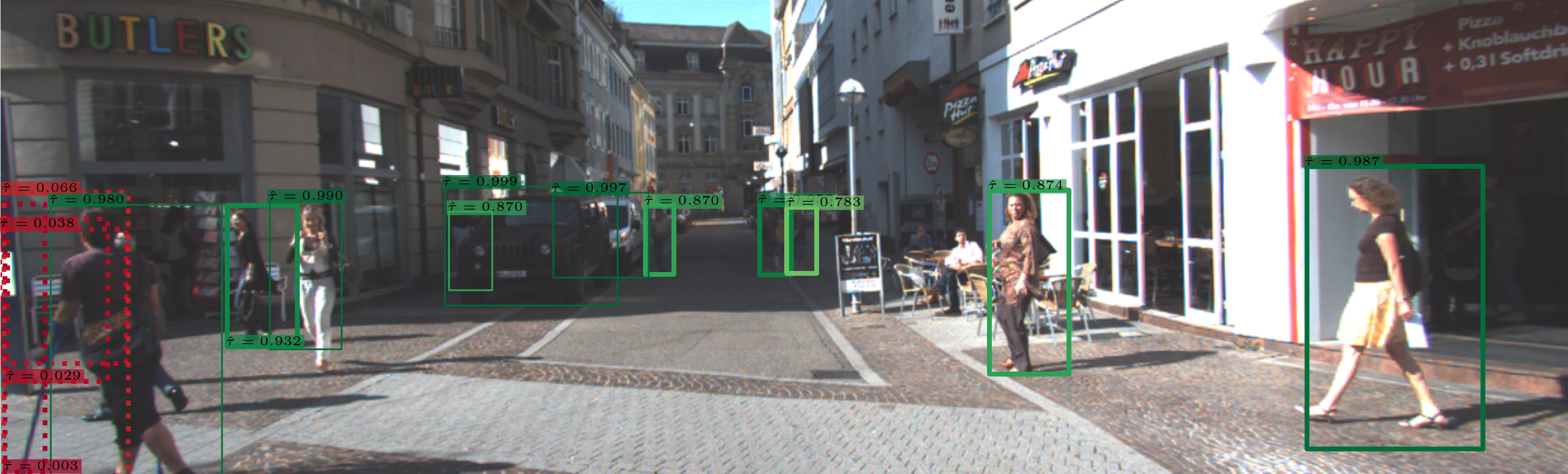}
    \caption{Full version of confidence estimation example in \cref{fig: reliability classifiers}.
    Top: DNN Score $\hat{s}$; bottom: meta classification confidence $\hat{\tau}$ involving gradient metrics.
    True predictions with low confidence are assigned large meta classification confidences while false predictions are assigned low values.
    This allows for improved filtering based on confidence.
    }
    \label{fig: full street scene example}
\end{figure*}

\subsection{Calibration.}
\label{app: calibration}
\newcommand{\TP}{\mathrm{TP}}
Generally, \enquote{calibration} methods (or re-calibration) aim at rectifying scores as confidences in the sense of \cref{sec: intro} such that the calibrated scores reflect the conditional frequency of true predictions.
For example, out of $100$ predictions with a confidence of $0.3$, around $30$ should be correct.

Confidence calibration methods have been applied to object detection in \cite{neumann2018relaxed} where temperature scaling was found to improve calibration. In addition to considering the expected calibration error (\(\ECE\)) and the maximum calibration error (\(\MCE\)) \cite{naeini2015obtaining_bayesian_binning}, the authors of \cite{neumann2018relaxed} argue that in object detection, it is important that confidences are calibrated irrespective of how many examples fall into a bin. Therefore, they introduced the average calibration error (\(\ACE\)) as a new calibration metric which is insensitive to the bin counts. The authors of \cite{kuppers2020multivariate} introduce natural extensions to localization-dependent calibration methods and a localization-dependent metric to measure calibration for different image regions.

In Sec.~\ref{sec: experiments}, we evaluated the calibration of meta classifiers in terms of the maximum (\(\MCE\), \cite{naeini2015obtaining_bayesian_binning}) and average (\(\ACE\), \cite{neumann2018relaxed}) calibration error which we define here.
We sort the examples into bins \(\beta_i\), \(i = 1, \ldots, B\) of a fixed width (in our case $0.1$, refer to \cref{sec: experiments} ) according to their confidence.
For each bin \(\beta_i\), we compute
\begin{align}
    \acc_i = \frac{\TP_i}{|\beta_i|}, \qquad
    \conf_i = \frac{1}{|\beta_i|} \sum_{j = 1}^{|\beta_i|} \hat c_i
\end{align}
where \(|\beta_i|\) denotes the number of examples in \(\beta_i\) and \(\hat c_i\) is the respective confidence, \ie, the network's score or a meta classification probability.
\(\TP_i\) denotes the number of correctly classified in \(\beta_i\).
In standard classification tasks, this boils down to the classification accuracy, whereas in the object detection setting, this is the detectors precision in the bin $\beta_i$.
A meta classifier performs a binary classification on detector positives, so we keep with the notation used for classifiers.
Calibration metrics are usually defined as functions of the bin-wise differences between $\mathrm{acc}_i$ and $\mathrm{conf}_i$.
In particular, we investigate the following calibration error metrics:
\begin{equation}
    \MCE =\, \max_{i = 1, \ldots, B}\, |\acc_i - \conf_i|,
\end{equation}
\begin{align}
    \ACE =&\, \frac{1}{B} \sum_{i = 1}^{B} |\acc_i - \conf_i|, \\
    \ECE =&\, \frac{1}{B} \sum_{i = 1}^{B} \frac{B}{|\beta_i|} |\acc_i - \conf_i|.
\end{align}
\Cref{tab: calibration apdx} also shows the expected (\(\ECE\), \cite{naeini2015obtaining_bayesian_binning}) calibration error
which was argued in \cite{neumann2018relaxed} to be biased toward bins with large amounts of examples.
\(\ECE\) is, thus, less informative for safety-critical investigations.

\section{Implemented Loss Functions}
\label{app: loss functions}
Here, we give a short account of the loss functions implemented in our experiments.

\paragraph{YOLOv3.}
The loss function we used to train YOLOv3 has the following terms:
\begin{align}
    \label{eq: mse yolo loss}
    \begin{split}
    \L_\xi^\mathrm{Yv3}(\hat{y},y)
    =& 2 \sum_{a = 1}^{N_\mathrm{out}}\sum_{t = 1}^{N_{\x}} \I_{at}^{\mathrm{obj}} \cdot  
    \left[\MSE\left(\begin{pmatrix}\widetilde \tau_\sfw^a \\ \widetilde \tau_\sfh^a \end{pmatrix}, \begin{pmatrix} \tau_\sfw^t \\ \tau_\sfh^t \end{pmatrix}\right)\right. \\ 
    &+ \left.\BCE\left(\sigma\begin{pmatrix} \widetilde \tau_\sfx^a \\ \widetilde \tau_\sfy^a \end{pmatrix}, \sigma \begin{pmatrix}\tau_\sfx^t \\ \tau_\sfy^t \end{pmatrix}\right)\right],
    \end{split}
    \\
    \label{eq: yolo score loss}
    \begin{split}
    \L_s^\mathrm{Yv3}(\widetilde{y},y)
    =& \sum_{a = 1}^{N_\mathrm{out}} \sum_{t = 1}^{N_{\x}} \left[\I_{at}^{\mathrm{obj}} \BCE_a\left(\sigma (\widetilde \tau_{s}), \mathbf{1}_{N_\mathrm{out}}\right)\right.
    \\
    &+ \left.\I_{at}^{\mathrm{noobj}}(y) \BCE_a\left(\sigma (\widetilde \tau_{s}), \mathbf{0}_{N_\mathrm{out}}\right)\right],
    \end{split}
    \\
    \L_p^\mathrm{Yv3}(\widetilde{y},y)
    =& \sum_{a = 1}^{N_\mathrm{out}} \sum_{t = 1}^{N_{\x}} \I_{at}^{\mathrm{obj}} \BCE\left(\sigma(\widetilde \tau_{p}^a), \sigma(\tau_{p}^t)\right).
\end{align}
Here, the first sum ranges over all \(N_{\out}\) anchors 
\(a\) and the second sum over the total number \(N_\mathrm{gt}\) of ground truth instances in \(y\).
\(\MSE\) is the usual mean squared error (\cref{eq: mse generic}) for the regression of the bounding box size.
As introduced in \cite{redmon2018yolov3}, the raw network outputs (in the notation of \cref{thm: complexity} this is one component \((\phi_T^{c})_{ab}\) of the final feature map \(\phi_T\), see also \cref{app: complexity}) for one anchor denoted by
\begin{equation}
    \label{eq: network output quantities}
    \widetilde{\tau} = (\widetilde \tau_{\mathsf{x}}, \widetilde \tau_{\mathsf{y}}, \widetilde \tau_{\mathsf{w}}, \widetilde \tau_{\mathsf{h}}, \widetilde \tau_s, \widetilde \tau_{p_1}, \ldots, \widetilde \tau_{p_C})
\end{equation}
are transformed to yield the components of \(\widetilde{y}\):
\begin{equation}
    \label{eq: transformation yolov3}
    \begin{split}
        \widetilde{\sfx} = \ell \cdot \sigma(\widetilde{\tau}_{\sfx}) + c_{\sfx}, \quad 
        \widetilde{\sfy} =& \,\ell \cdot \sigma(\widetilde{\tau}_{\sfy}) + c_{\sfy}, \quad 
        \widetilde{\sfw} = \pi_\sfw \cdot \mathrm{e}^{\widetilde{\tau}_\sfw}, \\
        \widetilde{\sfh} = \pi_\sfh \cdot \mathrm{e}^{\widetilde{\tau}_\sfh}, \quad
        \widetilde{s} =& \, \sigma(\widetilde{\tau}_s), \quad
        \widetilde{p}_j = \sigma(\widetilde{\tau}_{p_j}).
    \end{split}
\end{equation}
Here, \(\ell\) is the respective grid cell width/height, \(\sigma\) denotes the sigmoid function, \(c_\sfx\) and \(c_\sfy\) are the top left corner position of the respective cell and \(\pi_\sfw\) and \(\pi_\sfh\) denote the width and height of the bounding box prior (anchor).
These relationships can be (at least numerically) inverted to transform ground truth boxes to the scale of \(\widetilde{\tau}\).
We denote by \(\tau = (\tau_\sfx, \tau_\sfy, \tau_\sfw, \tau_\sfh, \tau_s, \tau_{p_1}, \ldots, \tau_{p_C})\) that transformation of the (real) ground truth \(y\) which is collected in a feature map \(\gamma\).
Then, we have
\begin{equation}
    \label{eq: mse generic}
    \MSE(\widetilde{\tau}, \tau) = \sum_{i} (\widetilde{\tau}_i - \tau_i)^2.
\end{equation}
Whenever summation indices are not clearly specified, we assume from the context that they take on all possible values, \eg in \cref{eq: mse yolo loss} \(\sfw\) and \(\sfh\).
Similarly, the binary cross entropy \(\BCE\) is related to the usual cross entropy loss \(\CE\) which is commonly used for learning probability distributions: 
\begin{align}
    \begin{split}
        \BCE(p, q) =&\, \sum_i \BCE_i(p, q) \\
        =&\, - \sum_i q_i \log(p_i) + (1 - q_i) \log(1 - p_i),
    \end{split}
        \\
        \CE(p, q) =&\, - \sum_i q_i \log(p_i)
\end{align}
where \(p, q \in (0, 1)^d\) for some fixed length \(d \in \N\).
Using binary cross entropy for classification amounts to learning \(C\) binary classifiers, in particular the \enquote{probabilities} \(\widetilde{p}_j\) are in general not normalized.
Note also, that each summand in \cref{eq: yolo score loss} only has one contribution due to the binary ground truth \(\mathbf{1}_{N_{\out}}\), resp.\ \(\mathbf{0}_{N_{\out}}\).
The binary cross entropy is also sometimes used for the center location of anchor boxes when the position within each cell is scaled to \((0, 1)\), see \(\L_\xi^{\mathrm{YOLOv3}}\).

The tensors \(\I^{\mathrm{obj}}\) and \(\I^{\mathrm{noobj}}\) indicate whether anchor \(a\) can be associated to ground truth instance \(t\) (obj) or not (noobj) which is determined as in \cite{ren2015faster} from two thresholds \(\varepsilon_+ \geq \varepsilon_- \geq 0\) which are set to 0.5 in our implementation.

Note, that both, \(\I^{\mathrm{obj}}\) and \(\I^{\mathrm{noobj}}\) only depend on the ground truth \(y\) and the fixed anchors, but not on the output regression results $\widetilde{y}$ or the predictions \(\hat{y}\).

We express a tensor with entries 1 with the size \(N\) as \(\mathbf{1}_N\) and a tensor with entries 0 as \(\mathbf{0}_N\).
The ground truth \(t\) one-hot class vector is \(\sigma(\tau_p^t) := p^t = (\delta_{i, \kappa_t})_{i = 1}^C\).

\paragraph{Faster R-CNN and Cascade R-CNN.}
Since Faster R-CNN \cite{ren2015faster} and Cascade R-CNN \cite{cai2018cascade} are two-stage architectures, there are separate loss contributions for the Region Proposal Network (RPN) and the Region of Interest (RoI) head, the latter of which produces the actual proposals.
Formally, writing \(\theta_\xi := (\theta_\sfx, \theta_\sfy, \theta_\sfw, \theta_\sfh)\) for the respectively transformed ground truth localization, similarly \(\widetilde{\theta}_\xi\) for the RPN outputs \(\widetilde{y}^\mathrm{RPN}\) and \(\widetilde{\theta}_s\) the proposal score output (where \(\widetilde{s}^a = \sigma(\widetilde{\theta}_s^a)\) is the proposal score):
\begin{align}
    \L_\xi^\mathrm{RPN}(\widetilde{y}^\mathrm{RPN}, y) 
    =&\, \frac{1}{|I^+|} \sum_{a = 1}^{N_\mathrm{out}^{\mathrm{RPN}}} \sum_{t = 1}^{N_{\x}} I^+_a \tilde\I_{at}^{\mathrm{obj}} \,\smL_\beta \left(\widetilde\theta_\xi^a, \theta_\xi^t\right), \\
    \begin{split}
        \L_s^\mathrm{RPN}(\widetilde{y}^\mathrm{RPN}, y)
    =&\, \sum_{a = 1}^{N_\mathrm{out}^{\mathrm{RPN}}} \sum_{t = 1}^{N_{\x}} 
        \left[            \I_{at}^{\mathrm{obj}} \BCE_a \left(\sigma(\widetilde{\theta}_s), \mathbf{1}_{N_\mathrm{out}^\mathrm{RPN}}\right)\right.\\
            &+\left.            I^-_a \tilde\I_{at}^{\mathrm{noobj}} \BCE_a\left(\sigma(\widetilde{\theta}_s), \mathbf{0}_{N_\mathrm{out}^\mathrm{RPN}}\right)        \right].
    \end{split}
\end{align}
The tensors \(\tilde\I^\mathrm{obj}\) and \(\tilde\I^\mathrm{noobj}\) are determined as for YOLOv3 with \(\varepsilon_+ = 0.7\) and \(\varepsilon_- = 0.3\)) but we omit their dependence on \(y\) in our notation.
Predictions are randomly sampled to contribute to the loss function by the tensors \(I^+\) and \(I^-\), which can be regarded as random variables.
The constant batch size \(b\) of predictions to enter the RPN loss is a hyperparameter set to $256$ in our implementation.
We randomly sample \(n_+ := \min\{|\tilde\I^\mathrm{obj}|, b/2\}\) of the \(|\tilde\I^\mathrm{obj}|\) positive anchors (constituting the mask \(I^+\)) and \(n_- := \min\{|\tilde\I^\mathrm{noobj}|, b - n_+\}\) negative anchors (\(I^-\)).
The summation of \(a\) ranges over the \(N_\mathrm{out}^\mathrm{RPN}\) outputs of the RPN (in our case, $1000$).
Proposal regression is done based on the smooth \(L^1\) loss 
\begin{equation}
    \smL_\beta (\widetilde{\theta}, \theta) := \sum_i
    \left\{    \begin{array}{l| l}
        \tfrac{1}{2} |\widetilde{\theta}_i - \theta_i|^2 & |\widetilde{\theta}_i - \theta_i| < \beta \\
        |\widetilde{\theta}_i - \theta_i| - \tfrac{\beta}{2} & |\widetilde{\theta}_i - \theta_i| \geq \beta
    \end{array}
    \right.,
\end{equation}
where we use the default parameter choice \(\beta = \tfrac{1}{9}\).
The final prediction \(\hat{y}\) of Faster R-CNN is computed from the proposals and RoI results\footnote{The result \(\widetilde{\tau}\) is similar to \cref{eq: network output quantities}, but without \(\widetilde{\tau}_s\) where we have instead \(C + 1\) classes, with one \enquote{background} class. We denote the respective probability by \(\widetilde p_0\).} \(\widetilde{\tau}\) in the RoI head:
\begin{align}
    \label{eq: transformation faster rcnn}
    \begin{split}
        \widetilde{\sfx} =&\, \pi_\sfw \cdot \widetilde{\tau}_\sfx + \pi_\sfx, \quad
        \widetilde{\sfy} = \pi_\sfy \cdot \widetilde{\tau}_\sfy + \pi_\sfy,\\
        \widetilde{\sfw} =&\, \pi_\sfw \cdot \e{\widetilde{\tau}_\sfw}, \quad
        \widetilde{\sfh} = \pi_\sfh \cdot \e{\widetilde{\tau}_\sfh}, \quad
        \widetilde{p} = \varSigma(\widetilde{\tau}_p),
    \end{split}
\end{align}
where \(\varSigma^i(x) := \e{x_i} / \sum_{j} \e{x_j}\) is the usual softmax function and \(\pi := (\pi_\sfx, \pi_\sfy, \pi_\sfw, \pi_\sfh)\) is the respective proposal localization.
Denoting with \(\tau_\xi := (\tau_\sfx, \tau_\sfy, \tau_\sfw, \tau_\sfh)\) ground truth localization transformed relatively to the respective proposal:
\begin{align}
    \L_\xi^\mathrm{RoI}(\widetilde{y}, y) =& \,
    \frac{1}{|\I^{\obj}|} \sum_{a = 1}^{N_\mathrm{out}} \sum_{t = 1}^{N_{\x}} \I_{at}^{\obj} \, \smL_\beta\left(\widetilde{\tau}_\xi^a, \tau_\xi^t\right), \\
    \begin{split}
        \L_p^\mathrm{RoI}(\widetilde{y}, y) =& \,
    \sum_{a = 1}^{N_{\out}} \sum_{t = 1}^{N_{\x}} \left[\I_{at}^{\obj} \, \CE(\varSigma(\widetilde{\tau}_p^a), p^t)\right.\\ 
    &+ \left.\I_{at}^{\noobj}\, \CE(\varSigma(\widetilde{\tau}_{p_0}^a), 1)\right].
    \end{split}
\end{align}
Here, \(\I^{\obj}\) and \(\I^{\noobj}\) are computed with \(\varepsilon_+ = \varepsilon_- = 0.5\).
The cascaded bounding box regression of Cascade R-CNN implements the smooth $L^1$ loss at each of three cascade stages, where bounding box offsets and scaling are computed from the previous bounding box regression results as proposals.

\paragraph{RetinaNet.}
In the RetinaNet \cite{lin2017focal} architecture, score assignment is part of the classification.
\begin{align}
    \L_\xi^\mathrm{Ret}(\widetilde{y}, y) =&\,
    \frac{1}{|\I^{\obj}|} \sum_{a = 1}^{N_{\out}} \sum_{t = 1}^{N_{\x}} \I^{\obj}_{at} L^1\left(\widetilde{\tau}_\xi^a, \tau_\xi^t\right), \\
    \begin{split}
    \L_p^\mathrm{Ret}(\widetilde{y}, y) =& \,
    \frac{1}{|\I^{\obj}|} \sum_{a = 1}^{N_{\out}} \sum_{t = 1}^{N_{\x}} \Big[        \I^{\obj}_{at} \,\sum_{j = 1}^C \alpha (1 - \sigma(\widetilde{\tau}_{p_j}^a))^{\gamma_\mathrm{F}} \cdot\\
        & \qquad\qquad\qquad\quad \cdot \BCE_j\left(\sigma(\widetilde{\tau}_p^a), p^t\right)\\
        + \I^{\noobj}_{at}& \, (1 - \alpha) \sigma(\widetilde{\tau}_{p_0}^a)^{\gamma_\mathrm{F}} \cdot \BCE\left(\sigma(\widetilde{\tau}_{p_0}^a), 0\right)    \Big].
    \end{split}
\end{align}
For \(\I^{\obj}\) and \(\I^{\noobj}\), we use \(\varepsilon_+ = 0.5\) and \(\varepsilon_- = 0.4\).
Regression is based on the absolute loss \(L^1(\widetilde{\tau}, \tau) = \sum_i |\widetilde{\tau}_i - \tau_i|\) and the classification loss is a formulation of the well-known focal loss with \(\alpha = 0.25\) and \(\gamma_\mathrm{F} = 2\).
Bounding box transformation follows the maps in \cref{eq: transformation faster rcnn}, where \(\pi\) are the respective RetinaNet anchor localizations instead of region proposals.
The class-wise scores of the prediction are obtained by \(\widetilde{p}_j = \sigma(\widetilde{\tau}_{p_j})\), \(j = 1, \ldots, C\) as for YOLOv3.

\paragraph{Theoretical loss derivatives.}
Here, we symbolically compute the loss gradients w.r.t.\ the network outputs as obtained from our accounts of the loss functions in the previous paragraphs.
We do so in order to determine the computational complexity for\(D_1 \L|_{\phi_T}\) in \cref{app: complexity}.
Note, that for all derivatives of the cross entropy, we can use
\begin{equation}
    \diff{}{\tau} \left[-y \log(\sigma(\tau)) - (1 - y) \log(1 - \sigma(\tau))\right]    = \sigma(\tau) - y.
\end{equation}
We then find for \(b = 1, \ldots, N_{\out}\) and features \(r \in \{\sfx, \sfy, \sfw, \sfh, s, p_1, \ldots, p_C\}\)
\begin{align}
    \dell{}{\widetilde\tau_r^b} \L^\mathrm{Yv3}_\xi =&\, 2\sum_{t = 1}^{N_{\x}}\I^{\obj}_{bt} \left\{    \begin{array}{l |l}
        \widetilde{\tau}_r^b - \tau_r^t & r \in \{\sfw, \sfh\} \\
        \sigma(\widetilde{\tau}_r^b) - \sigma(\tau_r^t) & r \in \{\sfx, \sfy\} \\
        0 & \mathrm{otherwise}.
    \end{array}
    \right., \\
    \dell{}{\widetilde{\tau}_r^b} \L^\mathrm{Yv3}_s =&\, \delta_{rs} \sum_{t = 1}^{N_{\x}} \left[        \I^{\obj}_{bt} (\widetilde{s}^b - 1) + \I^{\noobj}_{bt} \widetilde{s}^b
    \right], \\
    \dell{}{\widetilde{\tau}_r^b} \L^\mathrm{Yv3}_p =&\, \sum_{t = 1}^{N_{\x}} \I^{\obj}_{bt} \sum_{i = 1}^C \delta_{r p_i} (\widetilde{p}_i^b - p_i^t),
\end{align}
where \(\delta_{ij}\) is the Kronecker symbol, i.e., \(\delta_{ij} = 1\) if \(i = j\) and 0 otherwise.
Further, with analogous notation for the output variables of RPN and RoI
\begin{align}
    \begin{split}
        \dell{}{\hat\theta_r^b} \L_\xi^\mathrm{RPN} =&\, 
    \frac{1}{|I^+|} \sum_{t = 1}^{N_{\x}} I_b^+ \tilde \I_{bt}^{\obj}\, \cdot\\
    & \cdot\left\{    \begin{array}{l|l}
        \hat\theta_r^b - \theta_r^t & \begin{tabular}{@{}c@{}}$|\hat\theta_r^b - \theta_r^t| < \beta$ and \\ $r \in \{\sfx, \sfy, \sfw, \sfh\}$\end{tabular} \\
        \mathrm{sgn}(\hat\theta_r^b - \theta_r^t) & \begin{tabular}{@{}c@{}}$|\hat\theta_r^b - \theta_r^t| \geq \beta$ and \\ $r \in \{\sfx, \sfy, \sfw, \sfh\}$\end{tabular}\\
        0 & \mathrm{otherwise}
    \end{array}
    \right.,
    \end{split}
    \\
    \dell{}{\hat\theta_r^b} \L_s^\mathrm{RPN} =&\,
    \delta_{rs} 
    \left[        I_b^+ \tilde \I_{bt}^{\obj} (\hat{s}^b - 1) + I_b^- \tilde \I_{bt}^{\noobj} \hat{s}^b
    \right].
\end{align}
Here, \(\mathrm{sgn}\) denotes the sign function, which is the derivative of \(|\cdot|\) except for the origin.
Similarly,
\begin{align}
    \begin{split}
        \dell{}{\hat\tau_r^b} \L_\xi^\mathrm{RoI} =&\, 
    \frac{1}{|\I^{\obj}|} \sum_{t = 1}^{N_{\x}} \I_{bt}^{\obj} \, \cdot \\
    &\cdot\left\{    \begin{array}{l|l}
        \hat\tau_r^b - \tau_r^t & \begin{tabular}{@{}c@{}} $|\hat\tau_r^b - \tau_r^t| < \beta$ and\\$r \in \{\sfx, \sfy, \sfw, \sfh\}$ \end{tabular} \\
        \mathrm{sgn}(\hat\tau_r^b - \tau_r^t) & \begin{tabular}{@{}c@{}} $|\hat\tau_r^b - \tau_r^t| \geq \beta$ and\\ $r \in \{\sfx, \sfy, \sfw, \sfh\}$ \end{tabular}\\
        0 & \mathrm{otherwise}
    \end{array}
    \right.,
    \end{split}
    \\
    \begin{split}
        \dell{}{\hat{\tau}_r^b} \L_p^\mathrm{RoI} =&\,
    -\sum_{t = 1}^{N_{\x}} 
    \left[        \I_{bt}^{\obj} \sum_{j = 1}^C p_j^t \left(\delta_{p_j r} - \sum_{k = 0}^C \delta_{p_k r} \varSigma^k(\hat{\tau}_p^b)\right)\right.\\&+\left.\I^{\noobj}_{bt} \left(\delta_{p_0 r} - \sum_{k = 0}^C \delta_{p_k r} \varSigma^k(\hat{\tau}_p^b)\right)    \right].
    \end{split}
\end{align}
Note that the inner sum over \(j\) only has at most one term due to \(\delta_{p_j r}\).
With \(\sigma'(\tau) = \sigma(\tau) (1 - \sigma(\tau))\), we finally find for RetinaNet
\begin{align}
    \begin{split}
        \dell{}{\widetilde{\tau}_r^b} \L_\xi^\mathrm{Ret} &=\,
    \frac{1}{|\I^{\obj}|} \sum_{t = 1}^{N_{\x}} \I^{\obj}_{bt} \,\cdot \\
    &\cdot\left\{        \begin{array}{l|l}
            \mathrm{sgn}(\widetilde{\tau}_r^b - \tau_r^t) & r \in \{\sfx, \sfy, \sfw, \sfh\} \\
            0 & \mathrm{otherwise}
        \end{array}
    \right.,
    \end{split}
    \\
    \begin{split}
        \dell{}{\widetilde{\tau}_r^b} \L_p^\mathrm{Ret} &=\,
    \frac{1}{|\I^{\obj}|} \sum_{t = 1}^{N_{\x}} \Bigg[\I^{\obj}_{bt} \sum_{j = 1}^C \delta_{p_j r}
         \alpha (1 - \sigma(\widetilde{\tau}_{p_j}^b))^{\gamma_\mathrm{F}} \cdot \\ 
         &\cdot[-\gamma_\mathrm{F}\sigma(\widetilde{\tau}_{p_j}^b) \BCE_j(\sigma(\widetilde{\tau}_p^b), p^t) + \sigma(\widetilde{\tau}_{p_j}^b) - 1]
         \\\label{eq: derivative retinanet classification}
    &
    +
    \I^{\noobj}_{bt} \delta_{p_0 r} (1 - \alpha) \sigma(\widetilde{\tau}_{p_0}^b)^{\gamma_\mathrm{F}} \cdot \\ 
    &\cdot[-\gamma_\mathrm{F} (1 - \sigma(\widetilde{\tau}_{p_0}^b)) \log(1 - \sigma(\widetilde{\tau}_{p_0}^b)) + \sigma(\widetilde{\tau}_{p_0}^b)]
    \Bigg].
    \end{split}
\end{align}

\section{Implementation details}
\label{app: implementation details}
Here, we state details of the implementations of our framework to different architectures, and on different datasets.

\subsection{Datasets}
\label{app: datasets}
\begin{table}
    \caption{Dataset splits used for training and evaluation.}
    \centering
    \resizebox{\linewidth}{!}{
    \begin{tabular}{l c c c}
        \toprule
        Dataset & training & evaluation & \# eval images\\
        \midrule
        
        VOC  & 2007+2012 trainval & 2007 test & 4952 \\
        COCO  & train2017 & val2017 & 5000 \\
        
        KITTI  & \begin{tabular}{@{}c@{}} random part \\ of training\end{tabular} & \begin{tabular}{@{}c@{}} complement part \\ of training \end{tabular} & 2000 \\
        \bottomrule
    \end{tabular}
    }
    \label{tab: datasets}
\end{table}
In order to show a wide range of applications, we investigate our method on the following object detection datasets (see \cref{tab: datasets} for the splits used).

\paragraph{Pascal VOC 2007+2012 \cite{Everingham10}.}
The Pascal VOC dataset is an object detection benchmark of everyday images involving $20$ different object categories.
We train on the 2007 and 2012 trainval splits, accumulating to $16550$ train images and we evaluate on the 2007 test split of $4952$ images.
For training, we include labels marked as \enquote{difficult} in the original annotations.

\paragraph{MS COCO 2017 \cite{lin2014microsoft}.}
The MS COCO dataset constitutes a second vision benchmark involving 2D bounding box detection annotations for everyday images with $80$ object categories.
We train on the train2017 split of $118287$ images and evaluate on the $5000$ images of the val2017 split.

\paragraph{KITTI \cite{geiger2015kitti}.}
The KITTI vision benchmark contains $21$ real world street scenes annotated with 2D bounding boxes.
We randomly divide the $7481$ labeled images into a training split of $5481$ images and use the complement of $2000$ images for evaluation.

\subsection{Detectors}
\label{app: detectors}
For our experiments, we employ three common object detection architectures, namely YOLOv3 with Darknet53 backbone \cite{redmon2018yolov3}, Faster R-CNN \cite{ren2015faster} and RetinaNet \cite{lin2017focal}, each with a ResNet50FPN \cite{he2016deep} backbone.
Moreover, we investigate a state-of-the-art detector in Cascade R-CNN \cite{cai2018cascade} with a large ResNeSt200FPN \cite{zhang2020resnest} backbone.
We started from PyTorch \cite{NEURIPS2019_9015} reimplementations, added dropout layers and trained from scratch on the datasets in \cref{tab: datasets}.
We list some of the detector-specific details.

\paragraph{YOLOv3.}
The basis of our implementation is a publicly available GitHub repository \cite{westerndig2018}.
We position dropout layers with $p=0.5$ before the last convolutional layers of each detection head.
Gradient metrics are computed over the last two layers in each of the three detection heads as the final network layers have been found to be most informative in the classification setting \cite{oberdiek2018classification}.
Since each output box is the result of exactly one of the three heads, we only have two layers for gradients per box resulting in $2 \times 3$ gradients per box ($2$ layers per $3$ losses) as indicated in \cref{tab: number of layers losses and splits}.
We train an ensemble of $5$ detectors for each dataset from scratch.

\paragraph{Faster R-CNN.}
Based on the official Torchvision implementation, our model uses dropout ($p=0.5$) before the last fully connected layer of the architecture (classification and bounding box prediction in the Fast R-CNN head).
We compute gradient metrics for the last two fully connected layers of the Fast R-CNN head as well as for the last two convolutional layers of the RPN per box (objectness and localization), leading to $4 \times 2$ gradients per box ($2 + 2$ for localization, $2$ for classification and $2$ for proposal objectness).

\paragraph{RetinaNet.}
We also employ RetinaNet as implemented in Torchvision with ($p=0.5$)-dropout before the last convolutional layers for bounding box regression and classification.
Gradients are computed for the last two convolutional layers for bounding box regression and classification resulting in $2 \times 2$ gradients per prediction.

\paragraph{Cascade R-CNN.}
We use the Detectron2\cite{wu2019detectron2}-supported implementation of ResNeSt provided by the ResNeSt authors Zhang \etal \cite{zhang2020resnest} and the pre-trained weights on the MS COCO dataset.
We train from scratch on Pascal VOC and KITTI.
Since this model is primarily interesting for investigation due to its naturally strong score baseline based on cascaded regression, we do not report MC dropout results for it.
Gradient uncertainty metrics are computed for the last two fully connected layers (bounding box regression and classification) of each of the three cascades.
The loss of later cascade stages depends in principle on the weights of previous cascade stages. However, we only compute the gradients with respect to the weights in the current stage resulting in $2 \times 6$ ($3$ stages for bounding box regression and classification) gradients for the Cascade R-CNN head.
Furthermore, we have the $2 \times 2$ RPN gradients as in Faster R-CNN.

\subsection{Uncertainty baselines}
We give a short account of the baselines implemented and investigated in our experiments.

\paragraph{Score.} 
By the score, we mean the box-wise objectness score for YOLOv3 and the maximum softmax probability for Faster R-CNN, RetinaNet and Cascade R-CNN.
As standard object detection pipelines discard output bounding boxes based on a score threshold, this quantity is the baseline for discriminating true against false outputs.

\paragraph{Entropy.}
The entropy is a common \enquote{hand-crafted} uncertainty measure based on the classification output $\widetilde{p} \in[0, 1]^C$ (softmax or category-wise sigmoid) and given by 
\begin{equation}
    H(\widetilde{p}) = - \sum_{c = 1}^C \widetilde p_c \, \log(\widetilde p_c).
\end{equation}

\paragraph{Energy.}
\begin{table}
    \centering
    \caption{Ablation on the temperature parameter $T$ for the energy score in terms of meta classification ($\auroc$ and $\ap$) and meta regression ($R^2$).}
    \resizebox{\linewidth}{!}{
    \begin{tabular}{l c c c}\toprule
         & $\auroc$ & $\ap$ & $R^2$ \\\midrule
        $T = 1$ & $92.52 \pm 0.03$ & $91.86 \pm 0.04$ & $62.12 \pm 0.09$ \\
        $T = 10$ & $78.42 \pm 0.13$ & $81.75 \pm 0.08$ & $32.92 \pm 0.20$ \\
        $T = 100$ & $95.66 \pm 0.02$ & $95.33 \pm 0.03$ & $71.79 \pm 0.06$ \\
        $T = 1000$ & $95.62 \pm 0.03$ & $95.33 \pm 0.04$ & $71.78 \pm 0.05$ \\\midrule
        Score & $96.53 \pm 0.05$ & $96.87 \pm 0.03$ & $78.86 \pm 0.05$ \\
        MD & $\mathbf{98.23 \pm 0.02}$ & $\mathbf{98.06 \pm 0.02}$ & $\mathbf{85.88 \pm 0.10}$ \\
        $\mathrm{GS}_{\mathrm{full}}$ & \underline{$98.04 \pm 0.03$} & \underline{$97.81 \pm 0.06$} & \underline{$85.40 \pm 0.11$}\\\bottomrule
    \end{tabular}
    }
    \label{tab: ablation energy}
\end{table}
As an alternative to the maximum softmax probability and the entropy, Liu \etal proposed an energy score depending on a temperature parameter $T$ given by 
\begin{equation}
    E(\widetilde{\tau}) = - T \, \log \sum_{c = 1}^C \mathrm{e}^{\widetilde{\tau}_{p_c} / T}
\end{equation}
based on the probability logits $(\widetilde\tau_{p_1}, \ldots, \widetilde\tau_{p_C})$.
We found that $T = 100$ delivers the strongest results, see \cref{tab: ablation energy} where we compared different values of $T$ (like in \cite{liu2020energy}) for YOLOv3 on the KITTI dataset in terms of meta classification and meta regression performance.

\paragraph{Full softmax.}
We investigate an enveloping model of all classification-based uncertainty metrics by involving all probabilities $(\widetilde p_1, \ldots, \widetilde p_C)$ directly as co-variables in the meta classifier or meta regression model.
We find that it outperforms all purely classification-based models, which is expected.

\paragraph{MC dropout (MC).}
\begin{table}
    \centering
    \caption{Ablation on the sample count size $N_{\mathrm{MC}}$ for MC dropout in terms of meta classification ($\auroc$ and $\ap$) and meta regression ($R^2$).
    Results obtained from the sample standard deviation.
    }
    \resizebox{\linewidth}{!}{
    \begin{tabular}{l c c c}\toprule
         & $\auroc$ & $\ap$ & $R^2$ \\\midrule
        $N_{\mathrm{MC}} = 10$ & $97.40 \pm 0.04$ & $96.91 \pm 0.06$ & $80.85 \pm 0.10$ \\
        $N_{\mathrm{MC}} = 15$ & $97.50 \pm 0.03$ & $97.08 \pm 0.07$ & $81.28 \pm 0.09$ \\
        $N_{\mathrm{MC}} = 20$ & $97.69 \pm 0.03$ & $97.28 \pm 0.05$ & $82.11 \pm 0.09$ \\
        $N_{\mathrm{MC}} = 25$ & $97.64 \pm 0.03$ & $97.20 \pm 0.04$ & $81.94 \pm 0.12$ \\
        $N_{\mathrm{MC}} = 30$ & $97.60 \pm 0.07$ & $97.17 \pm 0.10$ & $82.10 \pm 0.11$ \\
        $N_{\mathrm{MC}} = 35$ & $97.71 \pm 0.03$ & $97.29 \pm 0.05$ & $82.17 \pm 0.13$ \\
        $N_{\mathrm{MC}} = 40$ & $97.69 \pm 0.04$ & $97.29 \pm 0.06$ & $82.12 \pm 0.13$ \\\midrule
        Score & $96.53 \pm 0.05$ & $96.87 \pm 0.03$ & $78.86 \pm 0.05$ \\
        MD & $\mathbf{98.23 \pm 0.02}$ & $\mathbf{98.06 \pm 0.02}$ & $\mathbf{85.88 \pm 0.10}$ \\
        $\mathrm{GS}_{\mathrm{full}}$ & \underline{$98.04 \pm 0.03$} & \underline{$97.81 \pm 0.06$} & \underline{$85.40 \pm 0.11$}\\\bottomrule
    \end{tabular}
    }
    \label{tab: ablation dropout}
\end{table}
As a common baseline, we investigate Monte-Carlo dropout uncertainty.
Since we are explicitly interested in the uncertainty content of MC dropout, we only include anchor-wise standard deviations of the entire network output $\widetilde{y}$ obtained from $30$ dropout samples.
We found that computing more samples does not significantly improve predictive uncertainty content as seen in the ablation study on the MC sample count $N_{\mathrm{MC}}$ in \cref{tab: ablation dropout} for YOLOv3 on the KITTI dataset.
Meta classification performance can be further improved by involving dropout means of $\widetilde{y}$.
However, MC dropout means do not carry an intrinsic meaning of uncertainty as opposed to standard deviations, so we do not include them in our main experiments.

\paragraph{Deep ensembles (E).}
\begin{table}
    \centering
    \caption{Ablation on the ensemble size $N_{\mathrm{ens}}$ for deep ensembles in terms of meta classification ($\auroc$ and $\ap$) and meta regression ($R^2$).
    Results obtained from the sample standard deviation.
    }
    \resizebox{\linewidth}{!}{
    \begin{tabular}{l c c c}\toprule
         & $\auroc$ & $\ap$ & $R^2$ \\\midrule
        $N_{\mathrm{ens}} = 3$ & $97.53 \pm 0.03$ & $97.17 \pm 0.05$ & $82.63 \pm 0.13$ \\
        $N_{\mathrm{ens}} = 4$ & $97.79 \pm 0.04$ & $97.48 \pm 0.06$ & $83.62 \pm 0.12$ \\
        $N_{\mathrm{ens}} = 5$ & $97.92 \pm 0.04$ & $97.63 \pm 0.05$ & $84.18 \pm 0.12$ \\
        $N_{\mathrm{ens}} = 6$ & $98.04 \pm 0.03$ & $97.75 \pm 0.04$ & $84.64 \pm 0.16$ \\
        $N_{\mathrm{ens}} = 7$ & $98.06 \pm 0.03$ & $97.80 \pm 0.05$ & $84.78 \pm 0.11$ \\
        $N_{\mathrm{ens}} = 8$ & \underline{$98.08 \pm 0.02$} & $97.80 \pm 0.03$ & $84.91 \pm 0.10$ \\\midrule
        Score & $96.53 \pm 0.05$ & $96.87 \pm 0.03$ & $78.86 \pm 0.05$ \\
        MD & $\mathbf{98.23 \pm 0.02}$ & $\mathbf{98.06 \pm 0.02}$ & $\mathbf{85.88 \pm 0.10}$ \\
        $\mathrm{GS}_{\mathrm{full}}$ & $98.04 \pm 0.03$ & \underline{$97.81 \pm 0.06$} & \underline{$85.40 \pm 0.11$}\\\bottomrule
    \end{tabular}
    }
    \label{tab: ablation ensemble}
\end{table}
As another common, sampling-based baseline, we investigate deep ensemble uncertainty obtained from ensembles of size $5$.
We find that larger ensembles do not significantly improve meta classification performance.
For reference, we show an ablation on the ensemble size $N_{\mathrm{ens}}$ for YOLOv3 on the KITTI dataset in \cref{tab: ablation ensemble} in terms of meta classification and meta regression.
By the same motivation like for MC dropout, we only include anchor-wise standard deviations over forward passes from the ensemble.

\paragraph{MetaDetect (MD).}
The output-based MetaDetect framework computes uncertainty metrics for use in meta classification and meta regression from pre-NMS variance in anchor-based object detection.
In our implementation, we compute the $46+C$ (where $C$ is the number of categories) MetaDetect metrics\cite{schubert2020metadetect} which include the entire network output $\widetilde{y}$.
The MetaDetect framework is, therefore, an enveloping model to any uncertainty metrics based on the object detection output (in particular to any classification-based uncertainty) which we also find in our experiments.
We include it in order to cover all such baselines.

\paragraph{Details of gradient-based uncertainty ($\mathrm{GS}$).}
In our experiments, we investigate two gradient-based uncertainty models.
While $\mathrm{GS}_{\norm{\cdot}_2}$ is based on the two-norms of box-wise gradients, $\mathrm{GS}_{\mathrm{full}}$ is utilizes all the six maps in \cref{eq: all maps for gradient metrics}.
While the two norms $\norm{\cdot}_1$ and $\norm{\cdot}_2$ directly compute the magnitude of a vector, the maps $\mean(\cdot)$ and $\std(\cdot)$ do not immediately capture a concept of length.
However, they have been found in \cite{oberdiek2018classification} to yield decent separation capabilities.
Similarly, the component-wise $\min(\cdot)$ and $\max(\cdot)$ contain relevant predictive information.
Note, that the latter two are related to the sup-norm $\norm{\cdot}_\infty$ but together contain more information.
While the last layer gradients themselves are highly informative, we allow for gradients of the last two layers in our main experiments.
In \cref{tab: ablation gradient depth} we show meta classification and meta regression performance of gradient-based models with metrics obtained from different numbers of network layers of the YOLOv3 model on the KITTI dataset.
Starting with the last layer gradient only (\# layers is 1), the gradient metrics from the two last layers and so on.
We see that meta classification performance quickly saturates and no significant benefit can be seen from using more than 3 layers.
However, meta regression can still be improved slightly  by using up to 5 network layers.

\begin{table*}
    \centering
    \caption{Ablation on the number of network layers used in terms of meta classification ($\auroc$ and $\ap$) and meta regression ($R^2$).
    Gradient metrics per layer are accumulated to those of later layers starting from the last layer of the DNN.}
    \resizebox*{0.9\textwidth}{!}{
    \begin{tabular}{c|c|ccccc}\toprule
        & & \multicolumn{5}{c}{\# layers} \\
        Metric & Score & 1 & 2 & 3 & 4 & 5 \\\midrule
        $\mathit{AuROC}$ & $96.53 \pm 0.05$ & $98.04 \pm 0.03$ & $98.06 \pm 0.02$ & $98.18 \pm 0.03$ & $98.18 \pm 0.03$ & $98.19 \pm 0.02$\\
        $\mathit{AP}$ & $96.87 \pm 0.03$ & $97.81 \pm 0.06$ & $97.83 \pm 0.04$ & $97.98 \pm 0.05$ & $98.00 \pm 0.04$ & $98.04 \pm 0.04$ \\
        $R^2$ & $78.89 \pm 0.05$ & $84.35 \pm 0.05$ & $85.40 \pm 0.11$ & $86.04 \pm 0.11$ & $86.18 \pm 0.07$ & $86.24 \pm 0.09$ \\\bottomrule
    \end{tabular}
    }
    \label{tab: ablation gradient depth}
\end{table*}

Throughout our experiments, we compute gradients via the PyTorch autograd framework, iteratively for each bounding box.
This procedure is computationally far less efficient than directly computing the gradients from the formulas in \cref{app: loss functions}.
We show in \cref{app: complexity}, that the latter is, in fact, at worst similar in FLOP count to MC dropout or deep sub-ensembles (an efficient implementation of deep ensembles).

In order to save computational effort, we compute gradient metrics not for all predicted bounding boxes.
We use a small score threshold of \(10^{-4}\) (KITTI, Pascal VOC), respectively \(10^{-2}\) (COCO) as a pre-filter.
On average, this produces \(\sim 150\) predictions per image.
These settings lead to a highly disbalanced TP/FP ratio post NMS on which meta classification and meta regression models are fitted.
On YOLOv3, for example, these ratios are for Pascal VOC: $0.099$, MS COCO: $0.158$ and for KITTI: $0.464$, so our models fit on significantly more FPs than TPs.
However, our meta classification and meta regression models (see \cref{sec: methods}) are gradient boosting models which tend to reflect well-calibrated confidences / regressions on the domain of training data.
Our results (\eg \cref{tab: meta classification performance}) obtained from cross-validation confirm that this ratio does not constitute an obstacle for obtaining well-performing models on data not used to fit the model.
For gradient boosting models, we employ the XGBoost library \cite{Chen:2016:XST:2939672.2939785} with $30$ estimators (otherwise standard settings).

\subsection{MetaFusion framework}
\label{app: metafusion}
\begin{figure}
    \centering
    \resizebox{\linewidth}{!}{
    \centering
\begin{tikzpicture}[scale=0.45, nodes=draw]
\node[ellipse, fill=gray!20] (0) at (-7, 0) {Detector};
\node[rectangle, fill=ao!20] (1) at (0, 1) {Bboxes};
\node[rectangle, fill=ao!20] (2) at (0, -0.5) {Score};
\node[rectangle, text width=1.8cm, align=center, fill=red!30] (3) at (-6, -3) {Uncertainty \\ Metrics};
\node[rectangle, text width=2cm, align=center, fill=red!30] (3a) at (0, -3) {Meta \\ Classification};
\node[ellipse, fill=ao!20] (4) at (7, 0) {Baseline};
\node[ellipse, fill=red!30] (5) at (7, -2) {MetaFusion};
\node[draw=none] (7) at (-3, 2) {NMS};
\node[draw=none] (7a) at (-3, -4.5) {};
\node[draw=none] (10) at (3.5, 2) {Threshold};
\node[draw=none] (10a) at (3.5, -4.5) {};

\draw[-stealth, out=0, in=180, ao] (0) to (1);
\draw[-stealth, out=0, in=180, ao] (0) to (2);
\draw[-stealth, out=-90, in=120, red] (0) to (3);
\draw[-stealth, out=0, in=180, red] (3) to (3a);
\draw (7) to (7a);
\draw (10) to (10a);
\draw[-stealth, in=180, out=0, ao] (1) to (4);
\draw[-stealth, in=180, out=0, ao] (2) to (4);
\draw[-stealth, in=180, out=0, red] (1) to (5);
\draw[-stealth, in=180, out=0, red] (3a) to (5);
\end{tikzpicture}
    }
    \caption{Schematic sketch of the baseline detection pipeline and the alternative MetaFusion pipeline for an object detector.}
    \label{fig: meta fusion pipeline}
\end{figure}
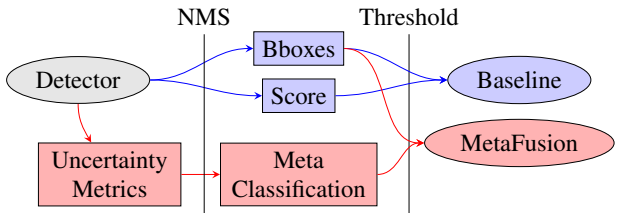
In \cref{sec: experiments}, we showed a way of trading uncertainty information for detection performance.
\Cref{fig: meta fusion pipeline} shows a sketch of the resulting pipeline, where the usual object detection pipeline is shown in blue.
The standard object detection pipeline relies on filtering out false positive output boxes on the basis of their score (see also \cref{fig: full street scene example}).
An altered confidence estimation like meta classification can improve the threshold-dependent detection quality of the object detection pipeline.
This way, boxes which are falsely assigned a low score can survive the thresholding step.
Similarly, FPs with a high score may be suppressed by proper predictive confidence estimation methods.
This approach is not limited to meta classification, however, our experiments show that meta classification constitutes such a method.

\section{Computational complexity}
\label{app: complexity}
In this section, we discuss the details of the setting in which \cref{thm: complexity} was formulated an give a proof for the statements made there.
The gradients for our uncertainty metrics are usually computed via backpropagation only for a few network layers.
Therefore, we restrict ourselves to the setting of fully convolutional neural networks.
\paragraph{Setting.}
As in \cite[Chapter 20.6]{shalev2014understanding}, we regard a (convolutional) neural network as a graph of feature maps with vertices \(V = \bigsqcup_{t = 0}^T V_t\) arranged in layers \(V_t\).
For our consideration it will suffice to regard them as sequentially ordered.
We denote \([n] := \{1, \ldots, n\}\) for \(n \in \N\).
Each layer \(V_t\) contains a set number \(k_t := |V_t|\) of feature map activations (channels) \(\phi_t^c \in \R^{h_t \times w_t}\), \(c \in[k_t]\).
We denote the activation of \(V_t\) by \(\phi_t = (\phi_t^1, \ldots, \phi_t^{k_t})\).
The activation \(\phi_{t + 1} \in (\R^{h_{t + 1} \times w_{t + 1}})^{k_{t + 1}}\) is obtained from \(\phi_t\) by convolutions.
We have \(k_t \times k_{t + 1}\) quadratic filter matrices
\begin{equation}
    (K_{t + 1})_c^{d} \in \R^{(2 s_t + 1) \times (2 s_t + 1)}, \quad c \in[k_t]; \quad d \in[k_{t + 1}],
\end{equation}
where \(s_t\) is a (usually small) natural number, the spatial extent of the filter.
Also, we have respectively \(k_{t + 1}\) biases \(b_{t + 1}^d \in \R\), \(d \in[k_{t + 1}]\).
The convolution (actually in most implementations, the cross correlation) of \(K \in \R^{(2 s + 1) \times (2 s + 1)}\) and \(\phi \in \R^{h \times w}\) is defined as
\begin{equation}
    \label{eq: convolution simple}
	(K \ast \phi)_{ab} := \sum_{m, n = - s}^s K_{s + 1 + p, s + 1 + q} \phi_{a + p, b + q},	
\end{equation}
where $a = 1, \ldots, h$ and $b = 1, \ldots, w$.
This is, strictly speaking, only correct for convolutions with stride 1, although a closed form can be given for the more general case.
For our goals, we will use stride 1 to upper bound the FLOPs which comes with the simplification that the feature maps' sizes are conserved.
We then define
 \begin{equation}
    \label{eq: convolution feature map result}
 	\psi_{t + 1}^d = \sum_{c = 1}^{k_t} (K_{t + 1})_c^d \ast \phi_t^c + b_{t + 1}^d \mathbf{1}_{h_t \times w_t}, \quad d \in[k_{t + 1}].
 \end{equation}
  Finally, we apply activation functions \(\alpha_t: \R \to \R\) to each entry to obtain \(\phi_{t + 1} = \alpha_{t + 1} (\psi_{t + 1})\).
 In practice, \(\alpha_t\) is usually a ReLU activation, i.e., \(\alpha_t(x) = \max\{x, 0\}\) or a slight modification (\eg leaky ReLU) of it and we will treat the computational complexity of this operation later.
 We can then determine the computational expense of computing \(\psi_{t + 1}\) from \(\phi_t\).
 In the following, we will be interested in the linear convolution action
 \begin{align}
    \begin{split}
        &C^{K_t} : \R^{k_{t - 1} \times h_{t - 1} \times w_{t - 1}} \to \R^{k_t \times h_t \times w_t}, \\
 	    &(C^{K_t} \phi_{t - 1})^d_{ab} := \left(\sum_{c = 1}^{k_t} (K_{t})_c^d \ast \phi_t^c\right)_{ab},
    \end{split}
 \end{align}
 where \(d \in[k_t]\), \(a \in[h_t]\) and \(b \in[w_t]\).
 Note that \(C^{K}\) is also linear in \(K_{t}\).
 On the last layer feature map \(\phi_T\) we define the loss function \(\mathcal{L}: (\phi_T, \gamma) \mapsto \mathcal{L}(\phi_T, \gamma) \in \R\).
 Here, \(\gamma\) stands for the ground truth\footnote{The transformations are listed in \cref{app: loss functions} for the entries \(\tau\) of \(\phi_T\).} transformed to feature map size \(\R^{h_T \times w_T \times k_T}\).
 In order to make dependencies explicit, define the loss of the sub-net starting at layer \(t\) by \(\ell_t\), \ie,
 \begin{equation}
    \label{eq: subnet losses}
 	\ell_T(\phi_T) := \mathcal{L}(\phi_T, \gamma),  \qquad
 	\ell_{t - 1}(\phi_{t-1}) := \ell_t(\alpha_t(\psi_t)).
 \end{equation}
 Straight-forward calculations yield
 \begin{align}
    \label{eq: simple last layer gradient}
    \begin{split}
 	\nabla\!{}_{K_T} \mathcal{L} =& \nabla_{K_T} \left(\ell_T \circ \alpha_T \circ \psi_T(K_T) \right)\\
 	=& D_1 \mathcal{L}|_{\phi_T} \cdot D \alpha_T|_{\psi_T} \cdot \nabla_{K_T} \psi_T
 	\end{split}
 	\\
 	\label{eq: simple 2-last layer gradient}
 	\begin{split}
 	\nabla\!{}_{K_{T - 1}} \mathcal{L} =& \nabla_{K_{T - 1}} \left(\ell_T \circ \alpha_T \circ \psi_T \circ \alpha_{T -1} \circ \psi_{T - 1}(K_{T - 1})\right)\\
 	=& D_1 \L|_{\phi_T} \cdot D \alpha_T|_{\psi_T} \cdot C^{K_T} \,\cdot \\
 	&\cdot D \alpha_{T - 1}|_{\psi_{T - 1}} \cdot \nabla_{K_{T - 1}} \psi_{T - 1}.
 	\end{split}
 \end{align}
 Here, \(D\) denotes the total derivative (\(D_1\) for the first variable, resp.) and we have used the linearity of \(C^{K_T}\).
 Note, that in \cref{sec: methods}, we omitted the terms \(D\alpha_T|_{\psi_T}\) and \(D\alpha_{T - 1}|_{\psi_{T - 1}}\).
 We will come back to them later in the discussion.
 For the gradient metrics we present in this paper, each \(\widetilde{y}^j\) for which we compute gradients receives a binary mask \(\mu^j\) such that \(\mu^j \cdot \phi_T\) are the feature map representations of candidate boxes for \(\widetilde{y}^j\) (see \cref{app: object detection}).
 The scalar loss function then becomes \(\mathcal{L}(\mu_j \phi_T, \gamma^j)\) for the purposes of computing gradient uncertainty, where \(\gamma^j\) is \(\overline{y}^j\) in feature map representation.
 We address next, how this masking influences \cref{eq: simple last layer gradient}, \cref{eq: simple 2-last layer gradient} and the FLOP count of our method.
 
 \paragraph{Computing the mask.}
 The complexity of determining \(\mu^j\) (\ie, finding \(\mathrm{cand}[\widetilde{y}^j]\)) is the complexity of computing all mutual \(\iou\) values between \(\widetilde{y}^j\) and the \(n_T := h_T \cdot w_T \cdot k_T\) other predicted boxes.
 Computing the \(\iou\) of a box \(b_1 = (x_1^{\min}, y_1^{\min}, x_1^{\max}, y_1^{\max})\) and  \(b_2 = (x_2^{\min}, y_2^{\min}, x_2^{\max}, y_2^{\max})\) can be done in a few steps with an efficient method exploiting the fact that:
 \begin{align}
 	U =& A_1 + A_2 - I, \qquad
 	\mathit{IoU} = I / U,
 \end{align}
 where the computation of the intersection area \(I\) and the individual areas \(A_1\) and \(A_2\) can each be done in 3 FLOP, resulting in 12 FLOP per pair of boxes.
 Note that different localization constellations of \(b_1\) and \(b_2\) may result in slightly varying formulas for the computation of \(I\) but the constellation can be easily detemined by binary checks which we ignore computationally.
 Also, the additional check for the class and sufficient score will be ignored, so we have \(12 n_T\) FLOP per mask \(\mu^j\).
 Inserting the binary mask\footnote{See \cref{sec: methods}. The mask \(\mu^j\) selects the feature map representation of \(\mathrm{cand}[\widetilde{y}^j]\) out of \(\phi_T\).} \(\mu^j\) in \cref{eq: simple last layer gradient} and \cref{eq: simple 2-last layer gradient} leads to the replacement of \(D_1 \L|_{\phi_T} \cdot D \alpha_T|_{\psi_T}\) by \(D \L^j := D_1 \L(\cdot, \gamma^j)|_{\mu^j\phi_T} \cdot \mu^j \cdot D \alpha_T|_{\psi_T}\) for each relevant box \(\widetilde{y}^j\).
 
In \cref{tab: flop and eval counts} we have listed upper bounds on the number of FLOP and elementary function evaluations performed for the computation of \(D \L^j\) for the investigated loss functions.
The numbers were obtained from the explicit partial derivatives computed in \cref{app: loss functions}.
In principle, those formulas allow for every possible choice of \(b \in[N_{\out}]\) which is why all counts are proportional to it.
Practically, however, at most the \(|\mu^j|\) candidate boxes are relevant which need to be identified additionally as foreground or background for \(\widetilde{y}^j\) in a separate step involving an \(\iou\) computation between \(\widetilde{y}^j\) and the respective anchor.
The total count of candidate boxes in practice is on average not larger than \(\sim 30\).
When evaluating the formulas from \cref{app: loss functions} note, that there is only one ground truth box per gradient and we assume here, that one full forward pass has already been performed such that the majority of the appearing evaluations of elementary functions (sigmoids, exponentials, etc.) have been computed beforehand.
This is not the case for the RetinaNet classification loss \eqref{eq: derivative retinanet classification}.
In \cref{tab: flop and eval counts} we also list the additional post-processing cost for the output transformations (see \cref{app: loss functions}, \cref{eq: transformation yolov3,eq: transformation faster rcnn}) required for sampling-based uncertainty quantification like MC dropout or deep ensemble samples (\enquote{sampling pp}).
The latter are also proportional to \(N_{\out}\), but also to the number \(N_\mathrm{samp}\) of samples.
 
 \begin{table*}[t]
     \centering
     \caption{Upper bounds on FLOP and elementary function evaluations performed during the computation of \(D \L^j\) (all contributions) and post processing for sampling-based uncertainty quantification (sampling pp) for \(N_\mathrm{samp}\) inference samples.
     }
     \resizebox{0.85\textwidth}{!}{
     \begin{tabular}{l c c c}\toprule
          & YOLOv3 & Faster/Cascade R-CNN & RetinaNet \\\midrule
        \# FLOP \(D \L^j\) & \((9 + C) N_{\out}\) & \(10 N_{\out}^\mathrm{RPN} + (2 + 2C) N_{\out}\) & \((18+11C) N_{\out}\)\\ 
        \# FLOP sampling pp & \(8 N_{\out} N_\mathrm{samp}\) & \((9 + 2C) N_{\out} N_\mathrm{samp}\) & \(8 N_{\out} N_\mathrm{samp}\)\\
        \# evaluations \(D \L^j\) & \(0\) & \(0\) & \(2 (1 + C) N_{\out}\)\\
        \# evaluations sampling pp & \((5 + C) N_{\out} N_\mathrm{samp}\) & \((3 + C) N_{\out} N_\mathrm{samp}\) & \((3 + C) N_{\out} N_\mathrm{samp}\)
        \\\bottomrule
     \end{tabular}
     }
     \label{tab: flop and eval counts}
 \end{table*}
 
 \paragraph{Proof of \cref{thm: complexity}.}
 Before we begin the proof, we first re-state the claims of \cref{thm: complexity}.
 
 \begin{thm-non}
    The number of FLOP required to compute the last layer ($t = T$) gradient \(\nabla_{\!K_T} \L(\mu^j \phi_T(K_T), \gamma^j)\) is \(\mathcal{O}(k_T h w + k_T k_{T - 1}(2s_T+1)^4)\).
    Similarly, for earlier layers $t$, \ie, \(\nabla_{\!K_t} \L(\mu^j \phi_T(K_t), \gamma^j)\), we have \(\mathcal{O}(k_{t + 1} k_{t} + k_{t} k_{t - 1})\), provided that we have previously computed the gradient for the consecutive layer \(t + 1\).
Performing variational inference only on the last layer, \ie, \(\phi_{T - 1}\) requires \(\mathcal{O}(k_T k_{T - 1} h w)\) FLOP per sample.
\end{thm-non}
 
 Our implementations exclusively use stride 1 convolutions for the layers indicated in \cref{sec: experiments}, so \(w_T = w_{T - 1} = w_{T - 2} =: w\), respectively \(h_T = h_{T - 1} = h_{T - 2} =: h\).
 As before, we denote \(n_t := h w k_t\), and regard \(D \L^j\) as a \(1 \times n_T\) matrix.
 Next, regard the matrix-vector multiplication to be performed in \cref{eq: simple last layer gradient}.
 Since for all \(t \in[T]\) we have that \(\psi_t\) is linear in \(K_t\), we regard \(\nabla_{\!K_t} \psi_t\) as a matrix acting on the filter space \(\R^{k_{t - 1} \times k_t \times (2 s_t + 1)^2}\).
 For \(d \in[k_t]\), \(\psi_t^d\) only depends on \(K_t^d\) (see \cref{eq: convolution feature map result}), so \(\nabla_{\!K_t} \psi_t\) only has at most \(k_{t - 1} \cdot (2 s_t + 1)^2 \cdot n_t\) non-vanishing entries.
 Therefore, regard it as a \((n_t \times (k_{t - 1} (2 s_t + 1)^2))\)-matrix.
 We will now show that this matrix has \(k_t (2 s_t + 1)^2\)-sparse columns.
 
 Let \(c \in[k_t]\), \(d \in[k_{t - 1}]\), \(p, q \in \{-s_t, \ldots, s_t\}\), \(a \in[h_t]\) and \(b \in[w_t]\).
 One easily sees from \cref{eq: convolution simple,eq: convolution feature map result} that
 \begin{equation}
     \dell{}{((K_{t})_c^d)_{pq}} (\psi_t)^d_{ab} = (\phi_{t - 1})^c_{a + p - s_t - 1, b + q - s_t - 1},
 \end{equation}
 where \(\phi_{t - 1}^c\) is considered to vanish for \(a + p - s_t -1 \notin[h_t]\) and \(b + q - s_t -1 \notin[w_t]\).
Consistency with the definition of \(p\) and \(q\) requires that both the conditions
\begin{equation}
    1 < a \leq 2 s_t + 2, \qquad 1 < b \leq 2 s_t + 2
\end{equation}
are satisfied, which means that \((\nabla_{\!K_t} \psi_t)^d\) can only have \(k_t (2 s_t + 1)^2\) non-zero entries.
Appealing to sparsity \(\nabla_{\!K_T} \psi_T\) in \cref{eq: simple last layer gradient} is then, effectively, a \(((k_{T - 1} \cdot (2 s_T + 1)^2) \times (k_T \cdot (2 s_T + 1)^2))\)-matrix, resulting in a FLOP count of 
\begin{equation}
    [2 \cdot k_T (2 s_T + 1)^2 - 1]	\cdot[k_{T - 1} \cdot (2 s_T + 1)^2]
\end{equation}
for the multiplication \(D \L^j \cdot \nabla_{\!K_T} \psi_T\) giving the claimed complexity considering that the computation of \(\mu^j\) is \(\mathcal{O}(k_T h w)\).

Next, we investigate the multiplication in \cref{eq: simple 2-last layer gradient}, in particular the multiplication \(D \L^j \cdot C^{K_{T}}\) as the same sparsity argument applies to \(\nabla_{\!K_{T - 1}} \psi_{T - 1}\).
First, for \(t \in[T]\), regard \(C^{K_t}\) as a \((n_t \times n_{t - 1})\)-matrix acting on a feature map \(\phi \in \R^{n_{t - 1}}\) from the left via
\begin{align}
    \begin{split}
        &\left(C^{K_t} \phi\right)_{ab}^d =
 	\sum_{c = 1}^{k_{t - 1}} \left[(K_t)_c^d \ast \phi^c\right]_{ab} \\
 	&\quad=\, \sum_{c = 1}^{k_{t - 1}} \sum_{m, n = - s_t}^{s_t} [(K_t)_c^d]_{s_t + 1 + m, s_t + 1 + n} (\phi^c)_{a + m, b + n},
    \end{split}
\end{align}
where \(d \in[k_t]\), \(b \in[w_t]\) and \(a \in[h_t]\) indicate one particular row in the matrix representation of \(C^{K_t}\).
From this, we see the sparsity of \(C^{K_t}\), namely the multiplication result of row \((d, a, b)\) acts on at most \(k_{t - 1} \cdot (2 s_t + 1)^2\) components of \(\phi_{t- 1}\) (i.e., \(k_{t - 1} (2 s_t + 1)^2\)-sparsity of the rows).
Conversely, we also see that at most \(k_t \cdot (2 s_t + 1)^2\) convolution products \((C^{K_t} \phi)_{ab}^d\) have a dependency on one particular feature map pixel \((\phi^c)_{\tilde{a} \tilde{b}}\) (\ie, \(k_t (2 s_t + 1)^2\)-sparsity of the columns).
Now, let \(t \in[T - 1]\) and assume that we have already computed the gradient
\begin{align}
    \begin{split}
        \nabla_{\!K_{t + 1}} \L
    =&\, \nabla_{\!K_{t + 1}} \ell_{t + 1}(\phi_{t + 1}(K_{t + 1}))\\
    =&\, D \ell_{t + 1}|_{\phi_{t + 1}} \cdot \alpha_{t + 1}|_{\psi_{t + 1}} \cdot \nabla_{\!K_{t + 1}} \psi_{t + 1},
    \end{split}
\end{align}
then by backpropagation, \ie, \cref{eq: subnet losses}, we obtain
\begin{align}
    \begin{split}
        \nabla_{\!K_t} \L
        =&\, \nabla_{\!K_t} [\ell_{t + 1} \circ \alpha_{t + 1} \circ \psi_{t + 1} (\phi_t(K_t))] \\
        =&\,  D \ell_{t + 1}|_{\phi_{t + 1}} \cdot \alpha_{t + 1}|_{\psi_{t + 1}} \cdot C^{K_{t + 1}} \,\cdot \\
        &\qquad \cdot D \alpha_t|_{\psi_t} \cdot \nabla_{\!K_t} \psi_t.
    \end{split}
\end{align}
Here, the first two factors have already been computed, hence we obtain a FLOP count for subsequently computing \(\nabla_{\!K_t}\L\) of
\begin{align}
    \label{eq: flops induction}
    \begin{split}
      &[2 \cdot k_{t+1}(2 s_{t+1} + 1)^2 - 1]\cdot[k_{t}(2 s_{t} + 1)^2]\, +\\
     &\quad + [2 \cdot k_{t} ( 2 s_{t} + 1)^2 - 1]	\cdot[k_{t - 1} (2 s_{t} + 1)^2]
    \end{split}
\end{align}
via the backpropagation step from \(\nabla_{\!K_{t+1}} \L\).
The claim in \cref{thm: complexity} addressing \cref{eq: simple 2-last layer gradient}, follows for \(t = T - 1\) in \cref{eq: flops induction}.

Finally, we address the computational complexity for sampling-based uncertainty quantification methods with sampling on \(\phi_{T - 1}\).
This is applicable, \eg, for dropout on the last layer (as in our experiments) or a deep sub-ensemble \cite{valdenegrodeep} sharing the forward pass up to the last layer (note, that we do not use sub-ensembles in our experiments, but regular deep ensembles).
Earlier sampling leads to far higher FLOP counts.
Again, we ignore the cost of dropout itself as it is random binary masking together with a respective up-scaling/multiplication of the non-masked entries by a constant.
The cost stated in \cref{thm: complexity} results from the residual forward pass \(\phi_{T - 1} \mapsto \phi_T = \alpha_T (C^{K_T} \cdot \phi_{T - 1} + b_T)\) where we now apply previous results.
Obtaining all \(n_T\) entries in the resulting sample featuOur meta classification and meta regression models (see \cref{sec: methods}) are gradient boosting models.
For gradient boosting models, we employ the XGBoost library \cite{Chen:2016:XST:2939672.2939785} with 30 estimators (otherwise standard settings).re map requires a total FLOP count of
\begin{equation}
    2 n_T k_{T - 1} (2 s_T + 1)^2 - 1 + n_T
\end{equation}
as claimed, where we have considered the sparsity of \(C^{K_T}\).
The last term results from the bias addition.

\paragraph{Discussion.}
A large part of the FLOP required to compute gradient metrics results from the computation of the masks \(\mu^j\) and the term \(D \L^j\) for each relevant predicted box.
In \cref{tab: flop and eval counts} we have treated the latter separately and found that, although the counts listed for \(D\L^j\) apply to each separate box, sampling post-processing comes with considerable computational complexity as well.
In that regard, we have similar costs for gradient metrics and sampling over the last network layer.
Note in particular, that computing \(D \L^j\) requires no new evaluation of elementary functions, as opposed to sampling.
Once \(D \L^j\) is computed for \(\widetilde{y}^j\), the last layer gradient can be computed in \(\mathcal{O}(k_T k_{T - 1})\) and every further gradient for layer \(V_t\) in \(\mathcal{O}(k_{t + 1} k_t + k_t k_{t - 1})\).
Each sample results in \(\mathcal{O}(n_T k_{T - 1})\) with sampling on \(\phi_{T - 1}\).
Sampling any earlier results in additional full convolution forward passes which also come with considerable computational costs.
We note that sampling-based epistemic uncertainty can be computed in parallel with all \(N_\mathrm{samp}\) forward passes being performed simultaneously.
Gradient uncertainty metrics, in contrast, require one full forward pass for the individual gradients \(\nabla_{\! K_t} \L(\mu^j \phi_T(K_t), \gamma^j)\) to be computed.
Therefore, gradient uncertainty metrics experience a slight computational latency as compared to sampling methods.
We argue that in principle, all following steps (computation of \(\mu^j\) and \(\nabla_{\!K_t} \L(\mu^j \phi_T(K_t), \gamma^j)\)) can be implemented to run in parallel as no sequential order of computations is required.
We have not addressed the computations of mapping the gradients to scalars from \cref{eq: all maps for gradient metrics} which are roughly comparable to the cost of computing the sample \(\std\) for sampling-based methods, especially once the sparsity of \(D\L^j\) has been determined in the computation of \(\nabla_{\!K_T} \L\).
The latter also brings a significant reduction in FLOP (from \(n_T\) to \(|\mu^j|\)) which cannot be estimated more sharply, however.
Since \(D\L^j\) is sparse, multiplication from the right with \(D\alpha_T|_{\psi_T}\) in \cref{eq: simple last layer gradient,eq: simple 2-last layer gradient} for a leaky ReLU activation only leads to lower-order terms.
The same terms were also omitted before in determining the computational complexity of sampling uncertainty methods.
Also, for this consideration, we regard the fully connected layers used for bounding box regression and classification in the Faster/Cascade R-CNN RoI head as \((1 \times 1)\)-convolutions to stay in the setting presented here.

\section{Further Numerical Results}
\label{app: results}

\subsection{Non-redundancy with output-based uncertainty}

\begin{table*}
    \centering
    \caption{
    Meta classification ($\auroc$ and $\ap$) and meta regression ($R^2$) performance of \colorbox{blue!70!green!10!white}{baseline methods}, \colorbox{red!10}{variants of gradient metrics} and \colorbox{orange!10}{different combinations of output-based uncertainty quantification methods} with gradient metrics ($\mean \pm \std$).
    We also show the results of using the entire network output $\widetilde{y}$ for meta classification and regression, as well, as adding sampling means to standard deviation features for MC and E.
    }
    \resizebox{\textwidth}{!}{
    \begin{tabular}{l | ccc  ccc  ccc}\toprule
         & \multicolumn{3}{c}{Pascal VOC} 
         & \multicolumn{3}{c}{COCO} 
         & \multicolumn{3}{c}{KITTI} \\\midrule
         \textbf{YOLOv3} & \(\auroc\) & \(\ap\) & \(R^2\)
         & \(\auroc\) & \(\ap\) & \(R^2\)
         & \(\auroc\) & \(\ap\) & \(R^2\) \\\midrule
         \rowcolor{blue!70!green!10!white}
        Score & $90.68 \pm 0.06$ & $69.56 \pm 0.12$ & $48.29 \pm 0.04$ & $82.97 \pm 0.04$ & $62.31 \pm 0.05$ & $32.60 \pm 0.02$ & $96.55 \pm 0.04$ & $96.87 \pm 0.03$ & $78.83 \pm 0.05$  \\
        \rowcolor{blue!70!green!10!white}
Entropy & $91.30 \pm 0.02$ & $61.94 \pm 0.06$ & $43.24 \pm 0.03$ & $76.52 \pm 0.02$ & $42.52 \pm 0.04$ & $21.10 \pm 0.04$ & $94.78 \pm 0.03$ & $94.82 \pm 0.05$ & $69.33 \pm 0.08$  \\
 \rowcolor{blue!70!green!10!white}
Energy & $92.59 \pm 0.02$ & $64.65 \pm 0.06$ & $47.18 \pm 0.03$ & $75.39 \pm 0.02$ & $39.72 \pm 0.06$ & $17.94 \pm 0.02$ & $95.46 \pm 0.05$ & $94.63 \pm 0.08$ & $70.39 \pm 0.10$  \\
 \rowcolor{blue!70!green!10!white}
Full Softmax & $93.81 \pm 0.06$ & $72.08 \pm 0.15$ & $53.86 \pm 0.11$ & $82.91 \pm 0.06$ & $58.65 \pm 0.10$ & $36.95 \pm 0.13$ & $97.10 \pm 0.02$ & $96.90 \pm 0.04$ & $78.79 \pm 0.12$  \\
 \rowcolor{blue!70!green!10!white}
Full output $\widetilde{y}$ & $95.84 \pm 0.04$ & $78.84 \pm 0.10$ & $60.67 \pm 0.18$ & $86.31 \pm 0.05$ & $67.46 \pm 0.07$ & $44.32 \pm 0.11$ & $98.35 \pm 0.02$ & $98.21 \pm 0.04$ & $86.34 \pm 0.07$  \\
 \rowcolor{blue!70!green!10!white}
MC${}_{\std}$ & $96.72 \pm 0.02$ & $78.15 \pm 0.09$ & $61.63 \pm 0.15$ & $89.04 \pm 0.02$ & $64.94 \pm 0.11$ & $43.85 \pm 0.09$ & $95.43 \pm 0.04$ & $94.11 \pm 0.12$ & $75.09 \pm 0.13$  \\
 \rowcolor{blue!70!green!10!white}
MC${}_{\std + \mean}$ & $97.42 \pm 0.02$ & $84.18 \pm 0.09$ & $68.33 \pm 0.16$ & $90.40 \pm 0.03$ & $72.63 \pm 0.07$ & $52.38 \pm 0.07$ & $98.43 \pm 0.03$ & $98.28 \pm 0.04$ & $86.86 \pm 0.09$  \\
 \rowcolor{blue!70!green!10!white}
E${}_{\std}$ & $96.87 \pm 0.02$ & $77.86 \pm 0.11$ & $61.48 \pm 0.07$ & $88.97 \pm 0.02$ & $64.05 \pm 0.12$ & $43.53 \pm 0.13$ & $97.98 \pm 0.03$ & $97.69 \pm 0.04$ & $84.29 \pm 0.12$  \\
 \rowcolor{blue!70!green!10!white}
E${}_{\std + \mean}$ & $97.62 \pm 0.02$ & $84.87 \pm 0.14$ & $68.88 \pm 0.09$ & $90.75 \pm 0.03$ & $73.15 \pm 0.06$ & $53.09 \pm 0.09$ & $98.61 \pm 0.02$ & \underline{$98.49 \pm 0.03$} & $88.00 \pm 0.08$  \\
\rowcolor{blue!70!green!10!white}
MC${}_{\std + \mean}$+E${}_{\std + \mean}$ & $97.69 \pm 0.02$ & $85.30 \pm 0.11$ & $69.60 \pm 0.13$ & $91.15 \pm 0.03$ & $73.85 \pm 0.05$ & $54.12 \pm 0.09$ & $98.61 \pm 0.01$ & \underline{$98.49 \pm 0.02$} & $87.95 \pm 0.10$ \\
 \rowcolor{blue!70!green!10!white}
MD & $95.78 \pm 0.05$ & $78.64 \pm 0.08$ & $60.36 \pm 0.14$ & $86.23 \pm 0.05$ & $67.37 \pm 0.08$ & $44.22 \pm 0.11$ & $98.23 \pm 0.03$ & $98.07 \pm 0.03$ & $85.97 \pm 0.09$  \\
 \rowcolor{red!10}
$\mathrm{GS}_{\norm{\cdot}_2}$ & $94.76 \pm 0.03$ & $74.86 \pm 0.10$ & $58.05 \pm 0.13$ & $84.90 \pm 0.02$ & $61.49 \pm 0.08$ & $38.77 \pm 0.04$ & $97.30 \pm 0.05$ & $96.82 \pm 0.10$ & $81.11 \pm 0.14$  \\
 \rowcolor{red!10}
$\mathrm{GS}_{\norm{\cdot}_{1, 2}}$ & $95.03 \pm 0.03$ & $76.04 \pm 0.10$ & $59.83 \pm 0.10$ & $86.21 \pm 0.04$ & $63.32 \pm 0.13$ & $41.36 \pm 0.09$ & $97.65 \pm 0.04$ & $97.21 \pm 0.07$ & $83.27 \pm 0.09$  \\
 \rowcolor{red!10}
$\mathrm{GS}_{\mathrm{full}}$ & $95.80 \pm 0.04$ & $78.57 \pm 0.11$ & $62.50 \pm 0.11$ & $86.94 \pm 0.04$ & $66.96 \pm 0.06$ & $44.90 \pm 0.09$ & $98.04 \pm 0.02$ & $97.81 \pm 0.04$ & $85.28 \pm 0.07$  \\
 \midrule
 \rowcolor{orange!10}
$\mathrm{GS}_{\mathrm{full}}$+$\widetilde{y}$ & $96.51 \pm 0.018$ & $81.20 \pm 0.09$ & $65.24 \pm 0.16$ & $87.54 \pm 0.04$ & $69.05 \pm 0.07$ & $47.67 \pm 0.09$ & $98.57 \pm 0.03$ & $98.47 \pm 0.04$ & $87.83 \pm 0.08$  \\
 \rowcolor{orange!10}
$\mathrm{GS}_{\mathrm{full}}$+MC${}_{\std}$ & $97.65 \pm 0.01$ & $85.12 \pm 0.06$ & $70.30 \pm 0.08$ & $90.76 \pm 0.02$ & $72.50 \pm 0.08$ & $52.71 \pm 0.07$ & $98.35 \pm 0.04$ & $98.16 \pm 0.04$ & $86.48 \pm 0.11$  \\
 \rowcolor{orange!10}
$\mathrm{GS}_{\mathrm{full}}$+E${}_\std$ & \underline{$97.85 \pm 0.02$} & \underline{$85.90 \pm 0.15$} & \underline{$71.22 \pm 0.07$} & \underline{$91.27 \pm 0.03$} & $73.44 \pm 0.06$ & \underline{$54.17 \pm 0.06$} & \underline{$98.64 \pm 0.02$} & \underline{$98.49 \pm 0.03$} & \underline{$88.34 \pm 0.10$}  \\
 \rowcolor{orange!10}
$\mathrm{GS}_{\mathrm{full}}$+MD & $96.46 \pm 0.04$ & $81.00 \pm 0.16$ & $65.08 \pm 0.14$ & $87.51 \pm 0.02$ & $68.98 \pm 0.08$ & $47.63 \pm 0.10$ & $98.53 \pm 0.03$ & $98.42 \pm 0.04$ & $87.69 \pm 0.06$  \\
 \midrule
 \rowcolor{blue!70!green!10!white}
MC${}_\std$+E${}_\std$+MD& $97.66 \pm 0.02$ & $85.13 \pm 0.12$ & $69.38 \pm 0.11$ & $91.14 \pm 0.02$ & \underline{$73.82 \pm 0.05$} & $54.07 \pm 0.08$ & $98.56 \pm 0.03$ & $98.45 \pm 0.03$ & $87.78 \pm 0.11$  \\
 \rowcolor{orange!10}
$\mathrm{GS}_{\mathrm{full}}$+MC${}_\std$+E${}_\std$+MD & $\mathbf{97.95 \pm 0.02}$ & $\mathbf{86.69 \pm 0.09}$ & $\mathbf{72.26 \pm 0.08}$ & $\mathbf{91.65 \pm 0.03}$ & $\mathbf{74.88 \pm 0.07}$ & $\mathbf{56.14 \pm 0.11}$ & $\mathbf{98.74 \pm 0.02}$ & $\mathbf{98.62 \pm 0.01}$ & $\mathbf{88.80 \pm 0.07}$  \\
        \bottomrule
    \end{tabular}
    }
    \label{tab: mc+mr vs baselines apdx}
\end{table*}

\begin{table*}
    \centering
    \caption{
    Meta classification ($\auroc$ and $\ap$) and meta regression ($R^2$) performance of \colorbox{blue!70!green!10!white}{baseline methods}, \colorbox{red!10}{variants of gradient metrics} and \colorbox{orange!10}{combinations of output- and gradient-based metrics} for different object detection architectures ($\mean \pm \std$).
    }
    \resizebox{\textwidth}{!}{
    \begin{tabular}{l | ccc  ccc  ccc}\toprule
         & \multicolumn{3}{c}{Pascal VOC} 
         & \multicolumn{3}{c}{COCO} 
         & \multicolumn{3}{c}{KITTI} \\\midrule
         & \(\auroc\) & \(\ap\) & \(R^2\)
         & \(\auroc\) & \(\ap\) & \(R^2\)
         & \(\auroc\) & \(\ap\) & \(R^2\) \\\midrule
\textbf{Faster R-CNN} &&&&&&&&& \\
\midrule
\rowcolor{blue!70!green!10!white}
Score & $89.77 \pm 0.05$ & $67.71 \pm 0.03$ & $39.94 \pm 0.02$ & $83.82 \pm 0.03$ & $64.14 \pm 0.03$ & $40.50 \pm 0.01$ & $96.53 \pm 0.05$ & $93.29 \pm 0.02$ & $72.29 \pm 0.02$  \\
 \rowcolor{blue!70!green!10!white}
MC & $89.99 \pm 0.06$ & $44.22 \pm 0.26$ & $23.70 \pm 0.17$ & $85.80 \pm 0.03$ & $40.48 \pm 0.12$ & $23.56 \pm 0.09$ & $93.39 \pm 0.07$ & $67.82 \pm 0.24$ & $40.09 \pm 0.17$  \\
 \rowcolor{blue!70!green!10!white}
MD & $94.43 \pm 0.02$ & $71.18 \pm 0.06$ & $47.92 \pm 0.09$ & $91.31 \pm 0.02$ & $64.73 \pm 0.05$ & $44.41 \pm 0.04$ & $98.86 \pm 0.03$ & $94.31 \pm 0.05$ & $79.92 \pm 0.04$  \\
 \rowcolor{red!10}
$\mathrm{GS}_{\norm{\cdot}_2}$ & $91.04 \pm 0.07$ & $61.66 \pm 0.15$ & $44.88 \pm 0.05$ & $89.80 \pm 0.03$ & $61.16 \pm 0.06$ & $44.93 \pm 0.04$ & $98.75 \pm 0.02$ & $93.01 \pm 0.05$ & $81.54 \pm 0.05$  \\
 \rowcolor{red!10}
$\mathrm{GS}_{\norm{\cdot}_{1, 2}}$ & $94.91 \pm 0.04$ & $67.73 \pm 0.10$ & $56.70 \pm 0.06$ & $90.64 \pm 0.03$ & $62.53 \pm 0.07$ & $48.27 \pm 0.03$ & $98.97 \pm 0.03$ & $93.89 \pm 0.07$ & $84.04 \pm 0.04$  \\
 \rowcolor{red!10}
$\mathrm{GS}_{\mathrm{full}}$ & $95.88 \pm 0.05$ & $68.74 \pm 0.13$ & $59.40 \pm 0.03$ & $91.38 \pm 0.03$ & $63.31 \pm 0.07$ & $50.44 \pm 0.04$ & $99.20 \pm 0.01$ & $94.60 \pm 0.07$ & $86.31 \pm 0.07$  \\
 \midrule
 \rowcolor{orange!10}
$\mathrm{GS}_{\mathrm{full}}$+MC & $96.59 \pm 0.03$ & $71.31 \pm 0.08$ & $60.74 \pm 0.07$ & $92.09 \pm 0.02$ & $64.59 \pm 0.06$ & $51.09 \pm 0.04$ & $99.34 \pm 0.02$ & $95.24 \pm 0.05$ & $86.85 \pm 0.04$  \\
 \rowcolor{orange!10}
$\mathrm{GS}_{\mathrm{full}}$+MD & \underline{$96.77 \pm 0.05$} & \underline{$73.60 \pm 0.07$} & \underline{$63.64 \pm 0.08$} & \underline{$92.30 \pm 0.02$} & \underline{$65.67 \pm 0.05$} & \underline{$52.30 \pm 0.04$} & \underline{$99.37 \pm 0.02$} & \underline{$95.38 \pm 0.05$} & \underline{$87.46 \pm 0.05$}  \\
 \rowcolor{orange!10}
$\mathrm{GS}_{\mathrm{full}}$+MC+MD & $\mathbf{96.72 \pm 0.04}$ & $\mathbf{73.51 \pm 0.10}$ & $\mathbf{63.02 \pm 0.03}$ & $\mathbf{92.30 \pm 0.01}$ & $\mathbf{65.77 \pm 0.06}$ & $\mathbf{52.21 \pm 0.04}$ & $\mathbf{99.35 \pm 0.02}$ & $\mathbf{95.37 \pm 0.03}$ & $\mathbf{86.99 \pm 0.07}$  \\
 \midrule
 \textbf{RetinaNet} &&&&&&&&& \\
 \midrule
 \rowcolor{blue!70!green!10!white}
 Score & $87.53 \pm 0.03$ & $66.30 \pm 0.05$ & $40.43 \pm 0.01$ & $84.95 \pm 0.04$ & $68.58 \pm 0.01$ & $39.88 \pm 0.02$ & $95.91 \pm 0.02$ & $89.93 \pm 0.02$ & $73.44 \pm 0.02$  \\
 \rowcolor{blue!70!green!10!white}
 MC & $72.90 \pm 0.08$ & $27.39 \pm 0.11$ & $14.17 \pm 0.12$ & $76.96 \pm 0.04$ & $43.54 \pm 0.06$ & $19.46 \pm 0.06$ & $88.13 \pm 0.06$ & $71.19 \pm 0.10$ & $50.51 \pm 0.12$  \\
 \rowcolor{blue!70!green!10!white}
MD & $89.57 \pm 0.04$ & $68.43 \pm 0.08$ & $50.27 \pm 0.10$ & $85.09 \pm 0.01$ & $68.32 \pm 0.06$ & $42.45 \pm 0.12$ & $96.19 \pm 0.03$ & $90.13 \pm 0.04$ & $77.53 \pm 0.08$  \\
 \rowcolor{red!10}
$\mathrm{GS}_{\norm{\cdot}_2}$ & $87.86 \pm 0.04$ & $64.35 \pm 0.06$ & $46.19 \pm 0.05$ & $81.62 \pm 0.04$ & $63.95 \pm 0.03$ & $38.01 \pm 0.04$ & $95.93 \pm 0.03$ & $90.03 \pm 0.05$ & $79.17 \pm 0.04$  \\
 \rowcolor{red!10}
$\mathrm{GS}_{\norm{\cdot}_{1, 2}}$ & $88.77 \pm 0.06$ & $65.40 \pm 0.05$ & $49.64 \pm 0.06$ & $83.53 \pm 0.05$ & $65.88 \pm 0.07$ & $41.96 \pm 0.05$ & $96.47 \pm 0.04$ & $90.50 \pm 0.03$ & $81.35 \pm 0.05$  \\
 \rowcolor{red!10}
$\mathrm{GS}_{\mathrm{full}}$ & $91.58 \pm 0.04$ & $68.32 \pm 0.06$ & $57.23 \pm 0.07$ & $85.59 \pm 0.02$ & $67.93 \pm 0.04$ & $47.74 \pm 0.06$ & $97.26 \pm 0.03$ & $91.51 \pm 0.07$ & $84.47 \pm 0.04$  \\
 \rowcolor{orange!10}
$\mathrm{GS}_{\mathrm{full}}$+MC & $92.54 \pm 0.03$ & $70.65 \pm 0.06$ & $61.73 \pm 0.04$ & $86.87 \pm 0.03$ & $69.42 \pm 0.03$ & $50.63 \pm 0.07$ & $97.52 \pm 0.02$ & $91.98 \pm 0.06$ & $85.08 \pm 0.04$  \\
 \rowcolor{orange!10}
$\mathrm{GS}_{\mathrm{full}}$+MD & $\mathbf{92.99 \pm 0.03}$ & \underline{$72.30 \pm 0.08$} & $\mathbf{64.32 \pm 0.07}$ & \underline{$87.15 \pm 0.05$} & \underline{$70.16 \pm 0.07$} & \underline{$51.07 \pm 0.09$} & \underline{$97.61 \pm 0.02$} & \underline{$92.26 \pm 0.05$} & $\mathbf{85.73 \pm 0.09}$  \\
 \rowcolor{orange!10}
$\mathrm{GS}_{\mathrm{full}}$+MC+MD & \underline{$92.95 \pm 0.03$} & $\mathbf{72.33 \pm 0.07}$ & \underline{$63.44 \pm 0.06$} & $\mathbf{87.20 \pm 0.04}$ & $\mathbf{70.21 \pm 0.03}$ & $\mathbf{51.38 \pm 0.09}$ & $\mathbf{97.63 \pm 0.01}$ & $\mathbf{92.30 \pm 0.03}$ & \underline{$85.64 \pm 0.08$}  \\
 \midrule
 \textbf{Cascade R-CNN} &&&&&&&&& \\
 \midrule
 \rowcolor{blue!70!green!10!white}
 Score & $95.70 \pm 0.04$ & $79.62 \pm 0.10$ & $57.90 \pm 0.09$ & $94.11 \pm 0.01$ & $81.36 \pm 0.02$ & $56.32 \pm 0.02$ & $98.67 \pm 0.02$ & $95.81 \pm 0.04$ & $83.31 \pm 0.03$  \\
 \rowcolor{blue!70!green!10!white}
 MD & $96.32 \pm 0.05$ & \underline{$82.11 \pm 0.12$} & $63.62 \pm 0.12$ & \underline{$94.12 \pm 0.03$} & \underline{$81.60 \pm 0.05$} & \underline{$58.84 \pm 0.04$} & $99.18 \pm 0.01$ & \underline{$96.60 \pm 0.05$} & $86.22 \pm 0.08$  \\
 \rowcolor{red!10}
$\mathrm{GS}_{\norm{\cdot}_{2}}$ & $96.46 \pm 0.05$ & $76.94 \pm 0.19$ & $61.56 \pm 0.12$ & $93.30 \pm 0.02$ & $76.40 \pm 0.06$ & $54.13 \pm 0.06$ & $99.19 \pm 0.01$ & $95.83 \pm 0.06$ & $85.80 \pm 0.06$  \\
 \rowcolor{red!10}
$\mathrm{GS}_{\norm{\cdot}_{1, 2}}$ & $96.54 \pm 0.06$ & $78.19 \pm 0.22$ & $62.82 \pm 0.15$ & $93.63 \pm 0.02$ & $77.95 \pm 0.06$ & $56.24 \pm 0.05$ & $99.23 \pm 0.01$ & $96.07 \pm 0.05$ & $86.33 \pm 0.06$  \\
 \rowcolor{red!10}
$\mathrm{GS}_{\mathrm{full}}$ & \underline{$96.66 \pm 0.05$} & $78.97 \pm 0.19$ & \underline{$63.94 \pm 0.13$} & $93.97 \pm 0.02$ & $79.17 \pm 0.09$ & $57.86 \pm 0.05$ & \underline{$99.34 \pm 0.01$} & $96.48 \pm 0.04$ & \underline{$87.39 \pm 0.08$}  \\
 \rowcolor{orange!10}
$\mathrm{GS}_{\mathrm{full}}$+MD & $\mathbf{97.24 \pm 0.05}$ & $\mathbf{84.11 \pm 0.13}$ & $\mathbf{69.78 \pm 0.13}$ & $\mathbf{94.78 \pm 0.02}$ & $\mathbf{82.53 \pm 0.05}$ & $\mathbf{62.13 \pm 0.05}$ & $\mathbf{99.48 \pm 0.01}$ & $\mathbf{97.27 \pm 0.04}$ & $\mathbf{89.59 \pm 0.04}$  \\
        \bottomrule
    \end{tabular}
    }
    \label{tab: mc+mr detectors apdx}
\end{table*}
Gradient metrics show significant improvements when combined with output- or sampling-based uncertainty quantification methods (see \cref{tab: meta classification performance} and \cref{tab: meta regression performance}).
We show additional meta classification and meta regression results in \cref{tab: mc+mr vs baselines apdx} and in \cref{tab: mc+mr detectors apdx} to further illustrate this finding.
First, in \cref{tab: mc+mr vs baselines apdx} we find that adding $\mathrm{GS}_{\mathrm{full}}$ to the raw object detection output features $\widetilde{y}$ performs similarly as the combination $\mathrm{GS}_{\mathrm{full}}$+MD.
In fact, when directly comparing MD with $\widetilde{y}$, we see consistently better results on $\widetilde{y}$, even though MD contains $\widetilde{y}$ as co-variables.
We attribute this finding to overfitting of the gradient boosting classifier and regression on MD.
This suggests that the information in MD is mostly redundant with the network output features.
Also, for combinations of one output-based uncertainty source (\ie, one of MC, E and MD) we gain strong boosts, especially in meta regression ($R^2$).
Note, that $\mathrm{GS}_{\mathrm{full}}$+E${}_\std$ is almost always the second-best model, even out-performing the purely output-based model MC${}_\std$+E${}_\std$+MD.
We show meta classification and meta regression performance of the sampling-based epistemic uncertainty methods MC and E when we include sampling averages of all features in addition to standard deviations which also leads to significant boosts.
Finally, we show an additional subset of $\mathrm{GS}_{\mathrm{full}}$ consisting of one- and two-norms ($\{\norm{\cdot}_1, \norm{\cdot}_2\}$) of all gradients which we abbreviate by $\mathrm{GS}_{\norm{\cdot}_{1,2}}$.
We notice significant gain of the latter to $\mathrm{GS}_{\norm{\cdot}_2}$, which shows that the one-norms $\norm{\cdot}_1$ contains important predictive information. Moreover, $\mathrm{GS}_{\mathrm{full}}$ is still significantly stronger than $\mathrm{GS}_{\norm{\cdot}_{1,2}}$, showing that the other uncertainty metrics in \cref{eq: all maps for gradient metrics} lead to large performance boosts.
Note that in almost all cases, combining MC dropout and deep ensemble metrics shows improvement over the single models even though both are epistemic (model) uncertainty.
The two methods, therefore, do not contain the exact same information but still complement each other to some degree and are rather different approximations of epistemic uncertainty

For further illustration of our method, \cref{tab: mc+mr detectors apdx} shows additional meta classification and meta regression results for the architectures from \cref{tab: meta performance networks}.
We find similar tendencies for the purely norm-based gradient model $\mathrm{GS}_{\norm{\cdot}_{1,2}}$ and see a significant degree of non-redundancy between gradient-based uncertainty and output-based uncertainty quantification methods.
Note in particular, that MC stays roughly on par with the score baseline in terms of $\auroc$.
We see significantly worse performance in terms of $\ap$ and meta regression ($R^2$).
We attribute this to the anchor-based dropout sampling method which was also employed for the present architectures (in the case of Faster R-CNN, the aggregation approach is proposal-based).

\begin{figure}
    \centering
    \includegraphics[width=\linewidth]{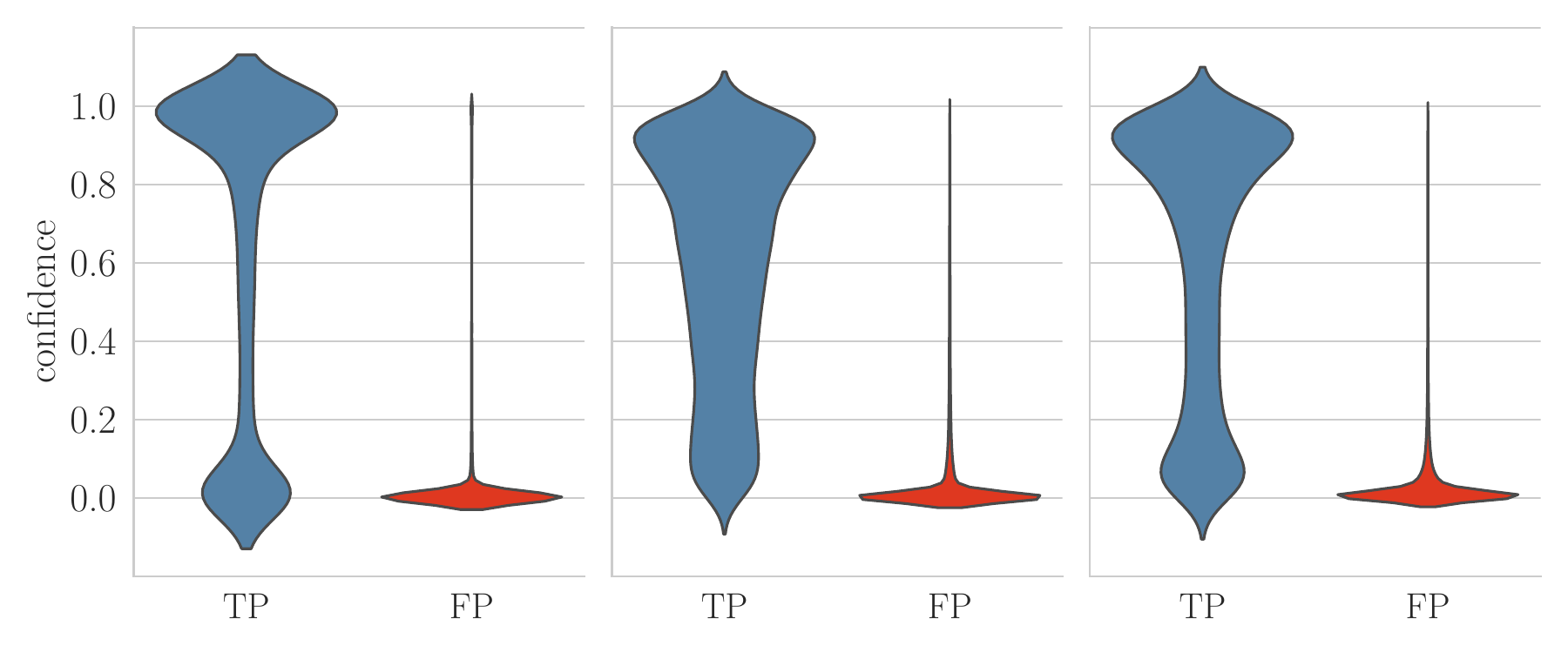}
    \caption{Confidence violin plots divided into TP and FP for Score (left), $\mathrm{GS}_{\mathrm{full}}$ (center) and $\mathrm{GS}_{\mathrm{full}}$+MC+E+MD (right).
    Model: YOLOv3, dataset: Pascal VOC evaluation split.
    }
    \label{fig: violins}
\end{figure}
\Cref{fig: violins} shows the confidence violin plots of the score (left), $\mathrm{GS}_{\mathrm{full}}$ (center) and  $\mathrm{GS}_{\mathrm{full}}$+MC+E+MD (right) conditioned on TP and FP predictions.
The violin widths are normalized for increased width contrast.
The score TP-violin shows especially large density at low confidences whereas the TP-violins of  $\mathrm{GS}_{\mathrm{full}}$ and  $\mathrm{GS}_{\mathrm{full}}$+MC+E+MD are less concentrated around the confidence $\hat \tau = 0$.
Instead, they have mass shifted towards the medium confidence range (\enquote{neck}).

\subsection{Calibration of meta classifiers}
\begin{figure*}
    \centering
    \includegraphics[width=\textwidth]{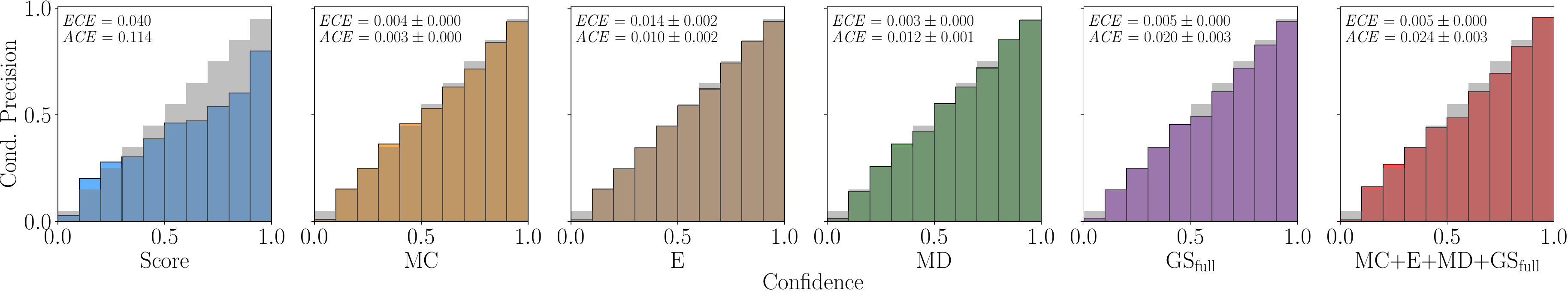}
    \caption{Reliability diagrams for the Score and meta classifiers based on different epistemic uncertainty metrics of the YOLOv3 architecture on the KITTI dataset.
    See \cref{tab: calibration apdx} for calibration errors of all meta classification models investigated in \cref{sec: experiments}.
    }
    \label{fig: apdx reliability}
\end{figure*}

\begin{table*}
    \centering
    \caption{Expected (\(\ECE\), \cite{naeini2015obtaining_bayesian_binning}), maximum (\(\MCE\), \cite{naeini2015obtaining_bayesian_binning}) and average (\(\ACE\), \cite{neumann2018relaxed}) calibration errors per confidence model over 10-fold cv (\(\mean \pm \std\)).}
    \resizebox{\textwidth}{!}{
    \begin{tabular}{l | ccc  ccc  ccc}\toprule
         & \multicolumn{3}{c}{Pascal VOC} 
         & \multicolumn{3}{c}{COCO} 
         & \multicolumn{3}{c}{KITTI} \\\midrule
         \textbf{YOLOv3} & \(\ECE\) & \(\MCE\) & \(\ACE\)
         & \(\ECE\) & \(\MCE\) & \(\ACE\)
         & \(\ECE\) & \(\MCE\) & \(\ACE\) \\\midrule
        \rowcolor{LightGray}
        Score & $0.040$ & $0.252$ & $0.114$ & $0.0327$ & $0.050$ & $0.034$ & $0.068$ & $0.348$ & $0.227$  \\
        Entropy & \underline{$0.002 \pm 0.001$} & \underline{$0.021 \pm 0.010$} & \underline{$0.007 \pm 0.003$} & $\mathbf{0.002 \pm 0.001}$ & $0.028 \pm 0.020$ & \underline{$0.007 \pm 0.003$} & $\mathbf{0.005 \pm 0.001}$ & $\mathbf{0.033 \pm 0.010}$ & $\mathbf{0.011 \pm 0.003}$ \\
        \rowcolor{LightGray}
        Energy Score & $\mathbf{0.001 \pm 0.001}$ & $\mathbf{0.015 \pm 0.007}$ & $\mathbf{0.005 \pm 0.002}$ & $\mathbf{0.002 \pm 0.001}$ & $0.021 \pm 0.003$ & $0.008 \pm 0.001$ & \underline{$0.006 \pm 0.002$} & \underline{$0.034 \pm 0.010$} & \underline{$0.013 \pm 0.005$} \\
        Full Softmax & $0.003 \pm 0.000$ & $0.028 \pm 0.006$ & $0.010 \pm 0.002$ & \underline{$0.003 \pm 0.001$} & \underline{$0.018 \pm 0.003$} & \underline{$0.007 \pm 0.001$} & $0.008 \pm 0.001$ & $0.048 \pm 0.010$ & $0.018 \pm 0.002$ \\
        \rowcolor{LightGray}
        MC & $0.004 \pm 0.000$ & $0.033 \pm 0.006$ & $0.014 \pm 0.002$ & $0.004 \pm 0.001$ & $0.025 \pm 0.003$ & $0.010 \pm 0.001$ & $0.011 \pm 0.001$ & $0.036 \pm 0.010$ & \underline{$0.013 \pm 0.002$} \\
        E & $0.003 \pm 0.000$ & $0.025 \pm 0.005$ & $0.010 \pm 0.002$ & $0.004 \pm 0.001$ & $0.022 \pm 0.003$ & $0.010 \pm 0.001$ & $0.013 \pm 0.001$ & $0.062 \pm 0.010$ & $0.028 \pm 0.004$ \\
        \rowcolor{LightGray}
        MD & $0.003 \pm 0.000$ & $0.040 \pm 0.009$ & $0.012 \pm 0.001$ & $0.005 \pm 0.001$ & $0.033 \pm 0.005$ & $0.014 \pm 0.001$ & $0.012 \pm 0.001$ & $0.074 \pm 0.020$ & $0.028 \pm 0.005$ \\
        $\mathrm{GS}_{\norm{\cdot}_2}$ & $0.003 \pm 0.000$ & $0.036 \pm 0.007$ & $0.014 \pm 0.001$ & $\mathbf{0.002 \pm 0.000}$ & $\mathbf{0.013 \pm 0.004}$ & $\mathbf{0.005 \pm 0.001}$ & $0.008 \pm 0.001$ & $0.054 \pm 0.010$ & $0.022 \pm 0.003$ \\
        \rowcolor{LightGray}
        $\mathrm{GS}_{\mathrm{full}}$ & $0.005 \pm 0.000$ & $0.055 \pm 0.002$ & $0.021 \pm 0.003$ & $0.005 \pm 0.001$ & $0.039 \pm 0.003$ & $0.015 \pm 0.001$ & $0.012 \pm 0.001$ & $0.078 \pm 0.020$ & $0.034 \pm 0.006$ \\\midrule
        MC+E+MD & $0.005 \pm 0.001$ & $0.049 \pm 0.010$ & $0.020 \pm 0.003$ & $0.005 \pm 0.000$ & $0.031 \pm 0.006$ & $0.014 \pm 0.001$ & $0.014 \pm 0.001$ & $0.076 \pm 0.010$ & $0.034 \pm 0.005$ \\
        MC+E+MD+$\mathrm{GS}_{\mathrm{full}}$ & $0.005 \pm 0.000$ & $0.061 \pm 0.010$ & $0.024 \pm 0.003$ & $0.006 \pm 0.000$ & $0.042 \pm 0.004$ & $0.018 \pm 0.001$ & $0.015 \pm 0.001$ & $0.106 \pm 0.020$ & $0.043 \pm 0.006$  \\
        \midrule
        \textbf{Faster R-CNN} &&&&&&&&& \\
        \midrule
        \rowcolor{LightGray}
        Score & $0.050$ & $0.427$ & $0.232$ & $0.075$ & $0.212$ & $0.138$ & $0.036$ & $0.283 $ & $0.114$  \\
        MD & $\mathbf{0.003 \pm 0.000}$ & $0.039 \pm 0.007$ & $0.013 \pm 0.002$ & $\mathbf{0.004 \pm 0.000}$ & $\mathbf{0.020 \pm 0.003}$ & $\mathbf{0.009 \pm 0.001}$ & $\mathbf{0.009 \pm 0.001}$ & $\mathbf{0.079 \pm 0.020}$ & $\mathbf{0.029 \pm 0.004}$  \\
        \rowcolor{LightGray}
        $\mathrm{GS}_{\mathrm{full}}$ & $0.004 \pm 0.000$ & $\mathbf{0.027 \pm 0.007}$ & $\mathbf{0.011 \pm 0.001}$ & $\mathbf{0.004 \pm 0.001}$ & $0.024 \pm 0.003$ & $\mathbf{0.009 \pm 0.001}$ & $0.010 \pm 0.001$ & $0.084 \pm 0.020$ & $0.035 \pm 0.004$  \\
        MD+$\mathrm{GS}_{\mathrm{full}}$ & $0.005 \pm 0.000$ & $0.044 \pm 0.007$ & $0.018 \pm 0.002$ & $0.006 \pm 0.001$ & $0.029 \pm 0.006$ & $0.012 \pm 0.001$ & $0.011 \pm 0.001$ & $0.088 \pm 0.010$ & $0.037 \pm 0.004$  \\
        \midrule
        \textbf{RetinaNet} &&&&&&&&& \\
        \midrule
        \rowcolor{LightGray}
        Score & $0.068$ & $0.212$ & $0.123$ & $0.089$ & $0.192$ & $0.106$ & $0.027$ & $0.097$ & $0.043$   \\
        MD & $\mathbf{0.003 \pm 0.000}$ & $\mathbf{0.031 \pm 0.008}$ & $\mathbf{0.011 \pm 0.002}$ & $\mathbf{0.005 \pm 0.001}$ & $\mathbf{0.022 \pm 0.004}$ & $\mathbf{0.009 \pm 0.001}$ & $\mathbf{0.003 \pm 0.000}$ & $\mathbf{0.044 \pm 0.006}$ & $\mathbf{0.016 \pm 0.002}$   \\
        \rowcolor{LightGray}
        $\mathrm{GS}_{\mathrm{full}}$ & $\mathbf{0.003 \pm 0.000}$ & $0.044 \pm 0.009$ & $0.014 \pm 0.001$ & $\mathbf{0.005 \pm 0.000}$ & $0.031 \pm 0.006$ & $0.012 \pm 0.001$ & $0.005 \pm 0.001$ & $0.060 \pm 0.010$ & $0.022 \pm 0.004$   \\
        MD+$\mathrm{GS}_{\mathrm{full}}$ & $0.005 \pm 0.000$ & $0.064 \pm 0.008$ & $0.024 \pm 0.002$ & $0.007 \pm 0.001$ & $0.032 \pm 0.004$ & $0.015 \pm 0.001$ & $0.006 \pm 0.000$ & $0.070 \pm 0.010$ & $0.028 \pm 0.003$   \\
        \midrule
        \textbf{Cascade R-CNN} &&&&&&&&& \\
        \midrule
        \rowcolor{LightGray}
        Score & $0.020$ & $0.219$ & $0.090$ & $0.029$ & $0.082$ & $0.042$ & $0.013$ & $0.188 $ & $0.078$   \\
        MD & $\mathbf{0.003 \pm 0.000}$ & $\mathbf{0.021 \pm 0.006}$ & $\mathbf{0.007 \pm 0.002}$ & $\mathbf{0.003 \pm 0.000}$ & $0.019 \pm 0.007$ & $\mathbf{0.006 \pm 0.001}$ & $\mathbf{0.002 \pm 0.000}$ & $\mathbf{0.038 \pm 0.010}$ & $\mathbf{0.016 \pm 0.005}$   \\
         \rowcolor{LightGray}
        $\mathrm{GS}_{\mathrm{full}}$ & $0.005 \pm 0.000$ & $0.032 \pm 0.010$ & $0.012 \pm 0.002$ & $\mathbf{0.003 \pm 0.000}$ & $\mathbf{0.017 \pm 0.003}$ & $0.007 \pm 0.001$ & $0.003 \pm 0.000$ & $0.052 \pm 0.010$ & $0.020 \pm 0.004$   \\
        MD+$\mathrm{GS}_{\mathrm{full}}$ & $0.005 \pm 0.000$ & $0.034 \pm 0.008$ & $0.014 \pm 0.002$ & $0.004 \pm 0.000$ & $0.025 \pm 0.004$ & $0.010 \pm 0.001$ & $0.003 \pm 0.000$ & $0.046 \pm 0.009$ & $0.019 \pm 0.003$   \\
        \bottomrule
    \end{tabular}
    }
    \label{tab: calibration apdx}
\end{table*}
For sake of completeness, we list in \cref{tab: calibration apdx} the calibration metrics \(\ECE\), \(\MCE\) and \(\ACE\) defined in \cref{app: object detection} for score and meta classifiers for all object detectors on all three datasets investigated in \cref{sec: experiments}.
All calibration metrics are in line with the results from \cref{sec: experiments} with meta classifiers being always better calibrated than the score by at least half an order of magnitude in any calibration metric.
See also \cref{fig: apdx reliability} for the additional reliability diagrams for MC, E and MC+E+MD$\mathrm{GS}_{\mathrm{full}}$ which extends \cref{fig: reliability example}.
The \(\ECE\) metric is comparatively small for all meta classifiers and, therefore, insensitive and harder to interpret than \(\MCE\) and \(\ACE\). 
As was argued in \cite{neumann2018relaxed}, the former is also less informative as bin-wise accuracy is weighted with the bin counts.
In \cref{tab: calibration apdx} we can see a weakly increasing trend of calibration errors in the meta classifiers due to overfitting on the increasing number of co-variables.
All meta classifiers are well-calibrated across the board.

\subsection{Meta regression scatter plots}
\begin{figure*}
    \centering
    \includegraphics[width=\linewidth]{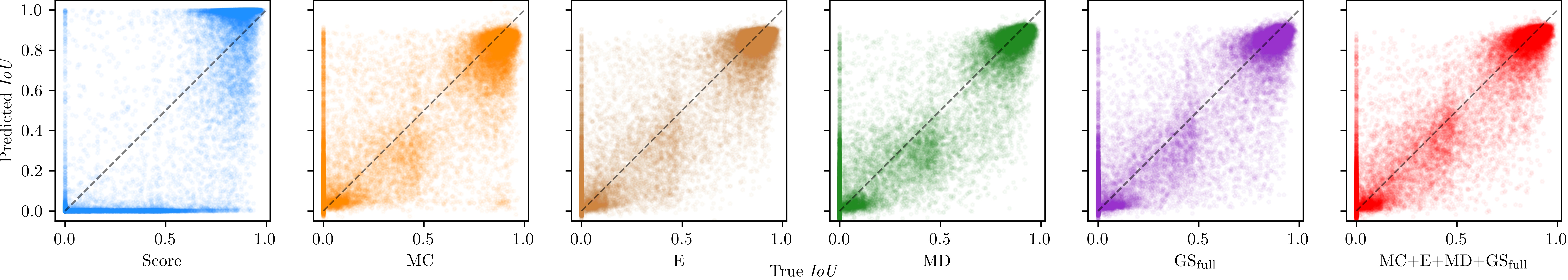}
    \caption{Scatter plots for samples of Score and meta regression based on MC dropout, gradient metrics G and the combination model G+MD+MC.
    We draw the optimal diagonal for reference.
    Model: YOLOv3, dataset: KITTI evaluation split.
    }
    \label{fig: meta regression scatter}
\end{figure*}
We underline the meta regression results obtained in \cref{sec: experiments} and \cref{app: results} by showing samples of predicted $\iou$ values over their true $\iou$ in \cref{fig: meta regression scatter}.
The samples are the results of one cross-validation split from \cref{tab: meta regression performance} and we indicate the diagonal of optimal regression with a dashed line in each panel.
Note that the $x$-axis shows the true $\iou$ values and we indicate the uncertainty quantification method below each panel plot at a label.
The $y$-axis shows the predicted $\iou$ for each method.
We find a large cluster for the score with low score but medium to high true $\iou$ (from $0.1$ to $0.8$), the right-most part of which (predicted $\iou \geq 0.5$) are false negative predictions.
In this regard, we refer again to \cref{fig: full street scene example} where FNs such as these become very apparent.
Moreover, the score indicates very little correlation with the true $\iou$ for true $\iou \geq 0.6$ where there are numerous samples with a score between $0.4$ and $0.6$.

In contrast, the meta regression models show striking amounts of FPs (true $\iou$ equal to $0$ and, \eg, prediction $\iota \geq 0.3$).
This phenomenon seems especially apparent for Monte Carlo dropout uncertainty.
The meta regression models MD,  $\mathrm{GS}_{\mathrm{full}}$ and  $\mathrm{GS}_{\mathrm{full}}$+MC+E+MD show fits that are comparatively close to the optimal diagonal which is in line with the determined regression performance $R^2$ between $0.81$ and $0.89$ in \cref{tab: meta regression performance}.

\subsection{MetaFusion on Pascal VOC.}
\begin{figure}
    \centering
    \includegraphics[width=\linewidth]{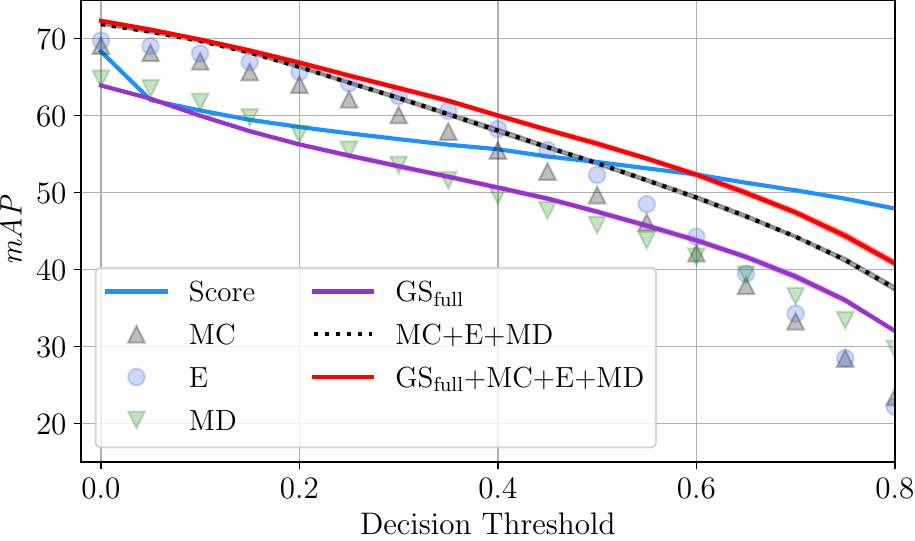}
    \caption{Score baseline and MetaFusion \(\map\) for the VOC evaluation dataset for YOLOv3 from 10-fold cv. 
    }
    \label{fig: fusion plots voc}
\end{figure}
\begin{table*}
    \centering
    \caption{COCO $\mathit{AP}$ metrics for the meta fusion models in \cref{fig: meta fusion map plot} at a confidence threshold of $\varepsilon_s = 0.05$.}
    \label{tab: ap fusion metrics}
    \resizebox{0.9\linewidth}{!}{
    \begin{tabular}{l | c c c c c c}
        \toprule
         & $\mathit{AP}$ & $\mathit{AP}_{50}$ & $\mathit{AP}_{75}$ & $\mathit{AP}_S$ & $\mathit{AP}_M$ & $\mathit{AP}_L$ \\
        \midrule
        \rowcolor{LightGray}
        Score & $58.9 $ & $88.25 $ & \underline{$68.27$} & $50.80$ & $60.00$ & $63.91$ \\
        MC & $57.58 \pm 0.18$ & $89.46 \pm 0.19$ & $65.55 \pm 0.30$ & $50.47 \pm 0.83$ & $59.12 \pm 0.14$ & $61.84 \pm 0.28$ \\
        \rowcolor{LightGray}
        E & \underline{$59.07 \pm 0.07$} & $90.40 \pm 0.07$ & $67.92 \pm 0.12$ & $52.37 \pm 0.51$ & \underline{$60.79 \pm 0.09$} & $63.15 \pm 0.17$ \\
        MD & $\mathbf{59.84 \pm 0.07}$ & $\mathbf{91.59 \pm 0.13}$ & $\mathbf{68.62 \pm 0.08}$ & \underline{$52.74 \pm 0.16$} & $\mathbf{60.95 \pm 0.08}$ & $\mathbf{64.72 \pm 0.20}$ \\
        \rowcolor{LightGray}
        $\mathrm{GS}_{\mathrm{full}}$ & $59.00 \pm 0.09$ & \underline{$90.65 \pm 0.12$} & $67.47 \pm 0.23$ & $\mathbf{53.53 \pm 0.16}$ & $59.80 \pm 0.14$ & \underline{$63.92 \pm 0.19$} \\
        \midrule
        MC+E+MD & $60.27 \pm 0.11$ & $91.91 \pm 0.10$ & $69.29 \pm 0.21$ & $53.33 \pm 0.32$ & $61.40 \pm 0.13$ & $64.99 \pm 0.15$ \\
        MC+E+MD+$\mathrm{GS}_{\mathrm{full}}$ & $60.35 \pm 0.09$ & $91.89 \pm 0.09$ & $69.41 \pm 0.18$ & $53.55 \pm 0.20$ & $61.43 \pm 0.14$ & $65.05 \pm 0.14$ \\
        \bottomrule
    \end{tabular}
    }
\end{table*}
In addition to the $\map$ plot in \cref{fig: meta fusion map plot} we show the COCO evaluation metrics $\mathit{AP}$, $\mathit{AP}_{50}$, $\mathit{AP}_{75}$, $\mathit{AP}_S$, $\mathit{AP}_M$, $\mathit{AP}_L$ in \cref{tab: ap fusion metrics} at a confidence threshold of $0.05$.
We see that MD performs strong across different metrics and the combined models consistently outperform the single metric models.
Gradient uncertainty models perform well for bounding boxes in the $S$ and $L$ categories with an improvement of $2.75$ ppts.\ over the Score baseline in$\mathit{AP}_S$.
Adding gradient uncertainty to MC+E+MD yields $0.22$ additional ppts.\ which is, however, within the 1 sigma overlap.
Note that the Score model is the second best in terms of$\mathit{AP}_{75}$, however, all meta classification models are fitted with a (meta) ground truth, where a prediction is classified as TP at the $\iou$ threshold $0.5$.
Fitting a meta classifiers at an $\iou$ threshold of $0.75$ is likely to perform better in the $\mathit{AP}_{75}$ metric.

We also show MetaFusion plots analogous to \cref{fig: meta fusion map plot} also for YOLOv3 on the  VOC2007 test split in \cref{fig: fusion plots voc}.
The qualitative behavior is similar to the one presented for the KITTI dataset in \cref{sec: experiments}, however, we do not see significant improvements by utilizing the purely gradient-based model $\mathrm{GS}_{\mathrm{full}}$.
Also, MC performs much stronger than on the KITTI dataset and is almost on par with E.
The combined models MC+E+MD and $\mathrm{GS}_{\mathrm{full}}$+MC+E+MD perform best, achieving higher maximum $\map$ than the score baseline by around $4$ ppts.
Note also, that there is a larger gap between MC+E+MD and $\mathrm{GS}_{\mathrm{full}}$+MC+E+MD for thresholds $\geq 0.2$ indicating gain from adding gradient uncertainty.
We point out that the score baseline shows a similar kink as in \cref{fig: meta fusion map plot} indicating a large amount of true predictions at low score values $\hat{s} \leq 0.05$.
\end{document}